\documentclass[10pt,twocolumn,twoside]{IEEEtran}

\usepackage{cite,graphicx,epstopdf,float,url,amssymb,amsthm,threeparttable,multirow,algorithmic,nameref,paralist}
\usepackage{amsfonts}
\usepackage{subcaption}
\usepackage[ruled,vlined]{algorithm2e}
\usepackage[usenames]{color}
\usepackage[normalem]{ulem}
\usepackage[cmex10]{amsmath}
\usepackage[keeplastbox]{flushend}
\IEEEoverridecommandlockouts

\usepackage{diagbox,array}
\newcolumntype{C}[1]{>{\centering\let\newline\\\arraybackslash\hspace{0pt}}m{#1}}
\newcolumntype{N}{@{}m{0pt}@{}}

\usepackage{etoolbox}%

\newcounter{numrel}%
\renewcommand{\thenumrel}{\alph{numrel}}%

\makeatletter
\newcommand{\numrel}[2]{%
  \refstepcounter{numrel}%
  \ltx@label{#2}%
  \overset{(\thenumrel)}{#1}%
}
\makeatother
\AfterEndEnvironment{IEEEproof}{\setcounter{numrel}{0}}%

\theoremstyle{plain}
\newtheorem{lemma}{Lemma}

\newtheorem{theorem}{Theorem}

\newtheorem{proposition}{Proposition}

\theoremstyle{definition}
\newtheorem{definition}{Definition}
\theoremstyle{remark}
\newtheorem{remark}{Remark}

\usepackage{url}
\usepackage[dvipsnames]{xcolor}
\usepackage[colorinlistoftodos,color=orange!70,textsize=scriptsize,shadow]{todonotes}
\usepackage{setspace}

\usepackage[normalem]{ulem}
\newcommand\redout{\bgroup\markoverwith{\textcolor{red}{\rule[.5ex]{2pt}{0.4pt}}}\ULon}

\newcommand{\mb}{\mathbf}
\newcommand{\mc}{\mathcal}

\newcommand{\mbb}{\mathbb}

\newcommand{\A}{\mb{A}}
\newcommand{\B}{\mb{B}}
\newcommand{\C}{\mb{C}}
\newcommand{\D}{\mb{D}}

\newcommand{\bH}{\mb{H}}
\newcommand{\I}{\mb{I}}
\newcommand{\bP}{\mb{P}}

\newcommand{\T}{\mb{T}}

\newcommand{\N}{\mb{W}}
\newcommand{\X}{\mb{X}}
\newcommand{\Y}{\mb{Y}}

\newcommand{\bd}{\mb{d}}

\newcommand{\bv}{\mb{v}}
\newcommand{\n}{\mb{w}}
\newcommand{\x}{\mb{x}}
\newcommand{\y}{\mb{y}}

\newcommand{\bbE}{\mbb{E}}
\newcommand{\bbR}{\mbb{R}}
\newcommand{\bbP}{\mbb{P}}

\newcommand{\cD}{\mc{D}}

\newcommand{\cJ}{\mc{J}}

\newcommand{\cS}{\mc{S}}
\newcommand{\cX}{\mc{X}}

\newcommand{\ul}{\underline}
\newcommand{\uN}{\ul{\N}}
\newcommand{\uX}{\ul{\X}}
\newcommand{\uY}{\ul{\Y}}

\newcommand{\wh}{\widehat}
\newcommand{\wt}{\widetilde}

\newcommand{\wD}{\wt{\D}}
\newcommand{\whD}{\wh{\D}}

\newcommand{\s}{\boldsymbol{\sigma}}
\newcommand{\eps}{\varepsilon}

\newcommand{\tet}{\boldsymbol{\theta}}

\newcommand{\vect}{\mathop{\mathrm{vec}}\nolimits}
\newcommand{\SNR}{\mathop{\mathrm{SNR}}\nolimits}
\newcommand{\supp}{\mathop{\mathrm{supp}}\nolimits}
\newcommand{\tr}{\mathop{\mathrm{Tr}}\nolimits}
\newcommand{\rnk}{\mathop{\mathrm{rank}}\nolimits}
\newcommand{\sign}{\mathop{\mathrm{sign}}\nolimits}
\newcommand{\RIP}{\mathsf{RIP}}

\newcommand{\norm}[1]{ \left\| #1 \right\| }

\newcommand{\dg}{\mathop{\mathrm{diag}}\nolimits}
\newcommand{\diag}[1]{ \dg\left( #1 \right) }
\newcommand{\Dg}{\mathop{\mathrm{Diag}}\nolimits}
\newcommand{\Diag}[1]{ \Dg\left( #1 \right) }

\newcommand{\lr}[1]{ \left\{ #1 \right\} }
\newcommand{\lrb}[1]{ \left[ #1 \right] }
\newcommand{\lrp}[1]{ \left( #1 \right) }
\newcommand{\lra}[1]{ \left| #1 \right| }

\newcommand{\bo}{\bigotimes}
\newcommand{\nonum}{\nonumber}

\newcommand{\Dks}{\D_{1:K}}
\newcommand{\Drks}{\D^0_{1:K}}
\newcommand{\Dkp}{\D'_{1:K}}

\newcommand{\Dhks}{\wh{\D}_{1:K}}
\newcommand{\rks}{\eps_{1:K}}

\newcommand{\boDk}{\bigotimes \D_k }
\newcommand{\boDrk}{\bigotimes \D^0_k }

\newcommand{\FnDk}{\|\D_k-\D'_k\|_F}
\newcommand{\FrDk}{\|\D_k-\D^0_k\|_F}

\newcommand{\PDk}{\bP_{\D_{k,\cJ_k}}}
\newcommand{\PD}{\bP_{\D_{\cJ}}}

\newcommand{\PDkp}{\bP_{\D'_{k,\cJ_k}}}
\newcommand{\PDrk}{\bP_{\D^0_{k,\cJ_k}}}

\newcommand{\HDk}{\bH_{\D_{k,\cJ_k}}}
\newcommand{\HD}{\bH_{\D_{\cJ}}}

\newcommand{\HDkp}{\bH_{\D'_{k,\cJ_k}}}
\newcommand{\HDrk}{\bH_{\D^0_{k,\cJ_k}}}

\newcommand{\PsDk}{\D_{k,\cJ_k}^+}
\newcommand{\PsD}{\D_{\cJ}^+}

\newcommand{\PsDkp}{{\D'}_{k,\cJ_k}^+}
\newcommand{\PsDrk}{{\D^0}_{k,\cJ_k}^+}

\IEEEoverridecommandlockouts

\begin{document}
\title{Identifiability of Kronecker-structured Dictionaries for Tensor Data}
\author{%
Zahra~Shakeri%
,
Anand~D.~Sarwate%
, and
Waheed~U.~Bajwa%
\thanks{This work is supported in part by the National Science Foundation under awards CCF-1525276 and CCF-1453073, and by the Army Research Office under award W911NF-17-1-0546. Some of the results reported here were presented at the 2017 IEEE International Workshop on Computational Advances in Multi-Sensor Adaptive Processing~\cite{shakeri2017tensor}.
}
\thanks{The authors are with the Department of Electrical and Computer Engineering, Rutgers, The State University of New Jersey, 94 Brett Rd, Piscataway, NJ 08854, USA (Emails: {\tt zahra.shakeri@rutgers.edu}, {\tt anand.sarwate@rutgers.edu}, and {\tt waheed.bajwa@rutgers.edu}).}
}

\maketitle

\begin{abstract}
This paper derives sufficient conditions for local recovery of coordinate dictionaries comprising a Kronecker-structured dictionary that is used for representing $K$th-order tensor data. Tensor observations are assumed to be generated from a Kronecker-structured dictionary multiplied by sparse coefficient tensors that follow the separable sparsity model. This work provides sufficient conditions on the underlying coordinate dictionaries, coefficient and noise distributions, and number of samples that guarantee recovery of the individual coordinate dictionaries up to a specified error, as a local minimum of the objective function, with high probability. In particular, the sample complexity to recover $K$ coordinate dictionaries with dimensions $m_k\times p_k$ up to estimation error $\eps_k$ is shown to be $\max_{k \in [K]}\mc{O}(m_kp_k^3\eps_k^{-2})$.
\end{abstract}


\section{Introduction}\label{sec:Introduction}
Rapid advances in sensing and data acquisition technologies are increasingly resulting in individual data samples or signals structured by multiple \textit{modes}. Examples include hyperspectral video (four modes; two spatial, one temporal, and one spectral), colored depth video (five modes; two spatial, one temporal, one spectral, and one depth), and four-dimensional tomography (four modes; three spatial and one temporal). Such data form multiway arrays and are called \textit{tensor data}~\cite{smilde2005multi,kolda2009tensor}.

Typical feature extraction approaches that handle tensor data tend to collapse or vectorize the tensor into a long one-dimensional vector and apply existing processing methods for one-dimensional data. Such approaches ignore the structure and inter-mode correlations in tensor data. More recently, several works instead assume a structure on the tensor of interest through tensor decompositions such as the CANDECOMP/PARAFAC (CP) decomposition~\cite{harshman1970foundations}, Tucker decomposition~\cite{tucker1963implications}, and PARATUCK decomposition~\cite{kolda2009tensor} to obtain meaningful representations of tensor data. Because these decompositions involve fewer parameters, or degrees of freedom, in the model, inference algorithms that exploit such decompositions often perform better than those that assume the tensors to be unstructured. Moreover, algorithms utilizing tensor decompositions tend to be more efficient in terms of storage and computational costs: the cost of storing the decomposition can be substantially lower, and numerical methods can exploit the structure by solving simpler subproblems.

In this work, we focus on the problem of finding sparse representations of tensors that admit a Tucker decomposition. More specifically, we analyze the \textit{dictionary learning} (DL) problem for tensor data. The traditional DL problem for vector-valued data involves constructing an overcomplete basis (dictionary) such that each data sample can be represented by only a few columns (atoms) of that basis~\cite{aharon2006img}. To account for the Tucker structure of tensor data, we require that the dictionary underlying the vectorized versions of tensor data samples be \textit{Kronecker structured} (KS). That is, it is comprised of \textit{coordinate dictionaries} that independently transform various modes of the tensor data. Such dictionaries have successfully been used for tensor data representation in applications such as hyperspectral imaging, video acquisition, distributed sensing, magnetic resonance imaging, and the tensor completion problem (multidimensional inpainting)~\cite{duarte2012kronecker,caiafa2013multidimensional}. To provide some insights into the usefulness of KS dictionaries for tensor data, consider the hypothetical problem of finding sparse representations of $1024 \times 1024 \times 32 $ hyperspectral images. Traditional DL methods require each image to be rearranged into a one-dimensional vector of length $2^{25}$ and then learn an unstructured dictionary that has a total of $(2^{25} p)$ unknown parameters, where $p \geq 2^{25}$. In contrast, KS DL only requires learning three coordinate dictionaries of dimensions $1024 \times p_1$, $1024 \times p_2$, and $32 \times p_3$, where $p_1,p_2\geq 1024$, and $p_3 \geq 32$. This gives rise to a total of $[1024 (p_1 + p_2) + 32p_3]$ unknown parameters in KS DL, which is significantly smaller than $2^{25} p$. While such ``parameter counting'' points to the usefulness of KS DL for tensor data, a fundamental question remains open in the literature: what are the theoretical limits on the learning of KS dictionaries underlying $K$th-order tensor data? To answer this question, we examine the KS-DL objective function and find sufficient conditions on the number of samples (or sample complexity) for successful local identification of \textit{coordinate dictionaries} underlying the KS dictionary. To the best of our knowledge, this is the first work presenting such identification results for the KS-DL problem.

\subsection{Our Contributions}\label{subsec:contr}
We derive sufficient conditions on the true coordinate dictionaries, coefficient and noise distributions, regularization parameter, and the number of data samples such that the KS-DL objective function has a local minimum within a small neighborhood of the true coordinate dictionaries with high probability. Specifically, suppose the observations are generated from a true dictionary $\D^0 \in \bbR^{m \times p}$ consisting of the Kronecker product of $K$ coordinate dictionaries, $\D_k^0 \in \bbR^{m_k \times p_k}, k \in \lr{1,\dots,K}$, where $m = \prod_{k=1}^Km_k$ and $p = \prod_{k=1}^Kp_k$. Our results imply that $N = \max_{k\in[K]} \Omega(m_kp_k^3\eps_k^{-2})$ samples are sufficient (with high probability) to recover the underlying coordinate dictionaries $\D_k^0$ up to the given estimation errors $\eps_k, k \in \lr{1,\dots,K}$.

\subsection{Relationship to Prior Work}\label{subsec:prior}

Among existing works on structured DL that have focused exclusively on the Tucker model for tensor data, several have only empirically established the superiority of KS DL in various settings for 2nd and 3rd-order tensor data~\cite{hawe2013separable,zubair2013tensor,caiafa2013multidimensional,roemer2014tensor,dantas2017learning,ghassemi2017stark}.

In the case of unstructured dictionaries, several works do provide analytical results for the dictionary identifiability problem~\cite{aharon2006uniqueness,agarwal2013learning,agarwal2013exact,arora2013new, schnass2014identifiability,schnass2014local,gribonval2014sparse,jung2015minimax}. These results, which differ from each other in terms of the distance metric used, cannot be trivially extended for the KS-DL problem. In this work, we focus on the Frobenius norm as the distance metric. Gribonval et al.~\cite{gribonval2014sparse} and Jung et al.~\cite{jung2015minimax} also consider this metric, with the latter work providing minimax lower bounds for dictionary reconstruction error. In particular, Jung et al.~\cite{jung2015minimax} show that the number of samples needed for reliable reconstruction (up to a prescribed mean squared error $\eps$) of an $m\times p$ dictionary within its local neighborhood must be \emph{at least} on the order of $N = \Omega(mp^2\eps^{-2})$. Gribonval et al.~\cite{gribonval2014sparse} derive a competing upper bound for the sample complexity of the DL problem and show that $N = \Omega(mp^3\eps^{-2})$ samples are \emph{sufficient} to guarantee (with high probability) the existence of a local minimum of the DL cost function within the $\eps$ neighborhood of the true dictionary. In our previous works, we have obtained lower bounds on the minimax risk of KS DL for 2nd-order~\cite{shakeri2016minimax} and $K$th-order tensors~\cite{shakeri2017sample,shakeri2016arxiv}, and have shown that the number of samples necessary for reconstruction of the true KS dictionary within its local neighborhood up to a given estimation error scales with the sum of the product of the dimensions of the coordinate dictionaries, i.e., $N = \Omega(p\sum_{k=1}^Km_kp_k\eps^{-2})$. Compared to this sample complexity lower bound, our upper bound is larger by a factor $\max_{k} p_k^2$.

In terms of the analytical approach, although we follow the same general proof strategy as the vectorized case of Gribonval et
al.~\cite{gribonval2014sparse}, our extension poses several technical challenges. These include: ($i$) expanding the asymptotic objective function into a summation in which individual terms depend on coordinate dictionary recovery errors, ($ii$) translating identification conditions on the KS dictionary to conditions on its coordinate dictionaries, and ($iii$) connecting the asymptotic objective function to the empirical objective function using concentration of measure arguments; this uses the \textit{coordinate-wise Lipschitz continuity} property of the KS-DL objective function with respect to the coordinate dictionaries. To address these challenges, we require additional assumption on the generative model. These include: ($i$) the true dictionary and the recovered dictionary belong to the class of KS dictionaries, and ($ii$) dictionary coefficient tensors follow the \textit{separable sparsity} model that requires nonzero coefficients to be grouped in blocks~\cite{caiafa2013computing,shakeri2016arxiv}.

\subsection{Notational Convention and Preliminaries} \label{subsec:notation}
Underlined bold upper-case, bold upper-case and lower-case letters are used to denote tensors, matrices and vectors, respectively, while non-bold lower-case letters denote scalars. For a tensor $\uX$, its $(i_1,\dots,i_K)$-th element is denoted as $\underline{x}_{i_1\dots i_K}$. The $i$-th element of vector $\mb{v}$ is denoted by $v_i$ and the $ij$-th element of matrix $\X$ is denoted as $x_{ij}$. The $k$-th column of $\X$ is denoted by $\x_k$ and $\X_{\mc{I}}$ denotes the matrix consisting of the columns of $\X$ with indices $\mc{I}$. We use $|\mc{I}|$ for the cardinality of the set $\mc{I}$. Sometimes we use matrices indexed by numbers, such as $\X_1$, in which case a second index (e.g., $\mb{x}_{1,k}$) is used to denote its columns. We use $\vect(\X)$ to denote the vectorized version of matrix $\X$, which is a column vector obtained by stacking the columns of $\X$ on top of one another. We use $\diag{\X}$ to denote the vector comprised of the diagonal elements of $\X$ and $\Diag{\bv}$ to denote the diagonal matrix, whose diagonal elements are comprised of elements of $\bv$. The elements of the sign vector of $\mb{v}$, denoted as $\sign(\mb{v})$, are equal to $\sign(v_i)= v_i/|v_i|$, for $v_i \neq 0$, and $\sign(v_i)=0$ for $v_i = 0$, where $i$ denotes the index of any element of $v$. We also use $\sin(\bv)$ to denote the vector with elements $\sin(v_i)$ (used similarly for other trigonometric functions). Norms are given by subscripts, so $\|\bv\|_0$, $\|\bv\|_1$, and $\|\bv\|_2$ are the $\ell_0$, $\ell_1$, and $\ell_2$ norms of $\mb{v}$, while $\|\X\|_2$ and $\|\X\|_F$ are the spectral and Frobenius norms of $\X$, respectively.
We use $[K]$ to denote $\{1,2,\dots,K\}$ and $\X_{1:K}$ to denote $\{\X_k\}_{k=1}^K$.

We write $\X \otimes \Y$ for the \textit{Kronecker product} of two matrices $\X\in \bbR^{m\times n}$ and $\Y\in \bbR^{p\times q}$,
where the result is an $mp \times nq$ matrix and we have $\|\X \otimes\Y \|_F = \|\X\|_F\| \Y\|_F$~\cite{horn2012matrix}. We also use $\bigotimes_{k \in K} \X_k \triangleq \X_1 \otimes \dots \otimes \X_K$ . We define $\bH_{\X}\triangleq (\X^\top \X)^{-1}$, $\X^+ \triangleq \bH_{\X}\X^\top$, and $\bP_{\X} \triangleq \X \X^+$ for full rank matrix $\X$. In the body, we sometimes also use $\Delta f(\X;\Y) \triangleq f(\X) - f(\Y)$.

For matrices $\X_1$ and $\X_2$ of appropriate dimensions, we define their distance to be $d(\X,\Y) = \|\X-\Y\|_F$. For $\X^0$ belonging to some set $\mc{X}$, we define
	\begin{align}
	\cS_{\eps}(\X^0) \triangleq \lr{\X \in \mc{X}: \|\X - \X^0\|_F = \eps}, \nonum \\
	\mc{B}_{\eps}(\X^0) \triangleq \lr{\X \in \mc{X}: \|\X - \X^0\|_F < \eps}, \nonum \\	
	\bar{\mc{B}}_{\eps}(\X^0) \triangleq \lr{\X \in \mc{X}: \|\X - \X^0\|_F \leq \eps}.
	\end{align}
Note that while $\cS_{\eps}(\X^0)$ represents the surface of a sphere, we use the term ``sphere" for simplicity. We use the standard ``big-$\mc{O}$'' (Knuth) notation for asymptotic scaling.

\subsubsection{Tensor Operations and Tucker Decomposition for Tensors}
A tensor is a multidimensional array where the order of the tensor is defined as the number of dimensions in the array.

\textit{Tensor Unfolding: }A tensor $\uX \in \bbR^{p_1 \times p_2\times \dots \times p_K}$ of order $K$ can be expressed as a matrix by reordering its elements to form a matrix. This reordering is called unfolding: the mode-$k$ unfolding matrix of a tensor is a $p_k \times \prod_{i \ne k} p_i$ matrix, which we denote by $\X_{(k)}$. Each column of $\X_{(k)}$ consists of the vector formed by fixing all indices of $\uX$ except the one in the $k$th-order.
The $k$-rank of a tensor $\uX$ is defined by $\rnk(\X_{(k)})$; trivially, $\rnk(\X_{(k)}) \leq p_k$.

\textit{Tensor Multiplication: } The mode-$k$ matrix product of the tensor $\uX$ and a matrix $\A \in \bbR^{m_k \times p_k}$, denoted by $\uX \times_k \A$, is a tensor of size $p_1 \times \dots p_{k-1} \times m_k \times p_{k+1} \dots \times p_K$ whose elements are
$
	(\uX \times_k \A)_{i_1\dots i_{k-1} j i_{k+1} \dots i_K} = \sum_{i_k=1}^{p_k} \underline{x}_{i_1\dots i_{k-1} i_k i_{k+1} \dots i_K} a_{ji_k}.
$
The mode-$k$ matrix product of $\uX$ and $\A$ and the matrix multiplication of $\X_{(k)}$ and $\A$ are related~\cite{kolda2009tensor}:
	\begin{align}
	\uY = \uX \times_k \A \Leftrightarrow \Y_{(k)} = \A \X_{(k)}.
	\end{align}
	
\textit{Tucker Decomposition: } The Tucker decomposition decomposes a tensor into a \textit{core tensor} multiplied by a matrix along each mode~\cite{tucker1963implications,kolda2009tensor}. We take advantage of the Tucker model since we can relate the Tucker decomposition to the Kronecker representation of tensors~\cite{caiafa2013computing}.
For a tensor $\uY \in \bbR^{m_1 \times m_2 \times \dots \times m_K}$ of order $K$, if $\rnk(\Y_{(k)})\leq p_k$ holds for all $k \in [K]$ then, according to the Tucker model, $\uY$ can be decomposed into:
	\begin{align} \label{eq:UY_UX}
	\uY = \uX \times_1 \D_1  \times_2 \D_2 \times_ 3 \dots \times_K \D_K,
	\end{align}
where $\uX \in \bbR^{p_1 \times p_2\times \dots \times p_K}$ denotes the core tensor and $\D_k \in \bbR^{m_k \times p_k}$ are factor matrices.
The following is implied by \eqref{eq:UY_UX}~\cite{kolda2009tensor}:
	\begin{align*}
	\Y_{(k)} = \D_{k}\X_{(k)}(\D_{K} \otimes \dots \otimes \D_{k+1} \otimes \D_{k-1} \otimes \dots \otimes \D_1)^\top.
	\end{align*}
Since the Kronecker product satisfies $\vect(\B\X\A^\top)=(\A \otimes \B)\vect(\X)$, \eqref{eq:UY_UX} is equivalent to
	\begin{align} \label{eq:vecty_vectx}
	\vect(\uY) = \big( \D_K \otimes \D_{K-1} \otimes \dots \otimes \D_1 \big) \vect(\uX),
	\end{align}
where $\vect(\uY) \triangleq \vect(\Y_{(1)})$ and $\vect(\uX) \triangleq \vect(\X_{(1)})$.

\subsubsection{Definitions for Matrices}
We use the following definitions for a matrix $\D$ with unit-norm columns:
$\delta_s(\D)$ denotes the \textit{restricted isometry property} ($\RIP$) constant of order $s$ for $\D$~\cite{candes2008restricted}.
We define the \textit{worst-case coherence} of $\D$ as $\mu_1(\D) = \max_{\substack{i,j\\i \neq j}} \lra{\bd_i^\top \bd_j}$.
We also define the \textit{order-$s$ cumulative coherence} of $\D$ as
	\begin{align} \label{eq:mu_1}
	\mu_{s}(\D) \triangleq \max_{|\cJ |\leq s} \max_{j \not\in \cJ}
		\|\D_{\cJ}^\top \bd_{j}\|_1.
	\end{align}
Note that for $s=1$, the cumulative coherence is equivalent to the worst-case coherence and $\mu_{s}(\D) \leq s \mu_1(\D)$~\cite{gribonval2014sparse}.
For $\D = \bo_{k \in [K]} \D_k$, where $\D_k$'s have unit-norm columns, $\mu_1(\D) = \max_{k \in [K]} \mu_1(\D_k)$~\cite[Corollary 3.6]{jokar2009sparse} and it can be shown that\footnote{The proof of \eqref{eq:mu_s} is provided in Appendix C.}:
	\begin{align}\label{eq:mu_s}
	\mu_s(\D) &\leq \max_{k \in [K]} \mu_{s_k}(\D_k)
		\bigg( \prod_{\substack{i \in [K], \\ i \neq k}}  \lrp{ 1+\mu_{s_i-1}(\D_i)} \bigg).
	\end{align}	

The rest of the paper is organized as follows. We formulate the KS-DL problem in Section~\ref{sec:model}. In Section~\ref{sec:asymp}, we provide analysis for asymptotic recovery of coordinate dictionaries composing the KS dictionary and in Section~\ref{sec:finite}, we present sample complexity results for identification of coordinate dictionaries that are based on the results of Section~\ref{sec:asymp}. Finally, we conclude the paper in Section~\ref{sec:discuss}. In order to keep the main exposition simple, proofs of the lemmas and propositions are relegated to appendices.

\section{System Model} \label{sec:model}
We assume the observations are $K$th-order tensors $\uY \in \bbR^{m_1\times m_2 \times \dots \times m_K}$. Given generating \textit{coordinate dictionaries} $\D^0_k \in \bbR^{m_k \times p_k}$, \textit{coefficient tensor} $\uX \in \bbR^{p_1\times p_2 \times \dots \times p_K}$, and \textit{noise tensor} $\uN$, we can write $\y \triangleq \vect(\uY)$ using \eqref{eq:vecty_vectx} as\footnote{We have reindexed $\D_k$'s in \eqref{eq:vecty_vectx} for ease of notation.}
	\begin{align} \label{eq:obs_model}
	\y = \bigg( \bo_{k \in [K]} \D_k^0 \bigg) \x + \n,
		\quad \|\x\|_0 \leq s,
	\end{align}
where $\x=\vect(\uX) \in \bbR^{p}$ denotes the sparse generating coefficient vector, $\D^0 = \bo \D_k^0 \in \bbR^{m\times p}$ denotes the underlying KS dictionary, and $\n=\vect(\uN)  \in \bbR^m$ denotes the underlying noise vector. Here, $\D_k^0 \in \mc{D}_k = \lr{ \D_k \in \bbR^{m_k\times p_k}, \|\bd_{k,j}\|_2 = 1, \forall j \in [p_k]}$ for $k \in [K]$, $p = \prod_{k \in [K]}p_k$ and $m = \prod_{k \in [K]}m_k$.\footnote{Note that the $\mc{D}_k$'s are compact sets on their respective oblique manifolds of matrices with unit-norm columns~\cite{gribonval2014sparse}.} We use $\bo$ for $\bo_{k\in[K]}$ in the following for simplicity of notation. We assume we are given $N$ noisy tensor observations, which are then stacked in a matrix $\Y = [\y_1,\dots,\y_N]$. To state the problem formally, we first make the following assumptions on distributions of $\x$ and $\n$ for each tensor observation.

\textit{Coefficient distribution:} We assume the coefficient tensor $\uX$ follows the random \textit{``separable sparsity"} model. That is, $\x=\vect(\uX)$ is sparse and the support of nonzero entries of $\x$ is structured and random. Specifically, we sample $s_k$ elements uniformly at random from $[p_k]$, $k \in [K]$. Then, the random support of $\x$ is $\lr{ \cJ \subseteq [p], |\cJ |=s}$ and is associated with
\begin{align*}
	\lr{\cJ_1\times \cJ_2 \times \dots \times \cJ_K:
		\cJ_k\subseteq [p_k], |\cJ_k|=s_k, k \in [K]}
\end{align*}
via lexicographic indexing, where $ s=\prod_{k \in [K]} s_k$, and the support of $\x_{1:N}$'s are assumed to be independent and identically distributed (i.i.d.). This model requires nonzero entries of the coefficient tensors to be grouped in blocks and the sparsity level associated with each coordinate dictionary to be small~\cite{caiafa2013computing}.\footnote{In contrast, for coefficients following the random non-separable sparsity model, the support of the nonzero entries of the coefficient vector are assumed uniformly distributed over $\lr{\cJ \subseteq [p]: |\cJ|=s}$.}

We now make the same assumptions for the distribution of $\x$ as assumptions A and B in Gribonval et al.~\cite{gribonval2014sparse}.
These include:
($i$) $\bbE\lr{\x_\cJ \x_\cJ^\top |\cJ} = \bbE\lr{x^2}\I_{s}$,
($ii$) $\bbE\lr{\x_\cJ \s_\cJ^\top |\cJ} = \bbE \lr{|x|}\I_{s}$, where $\s = \sign(\x)$,
($iii$) $\bbE\lr{\s_\cJ \s_\cJ^\top |\cJ} = \I_{s}$,
($iv$) magnitude of $\x$ is bounded, i.e., $\|\x\|_2 \leq M_x $ almost surely, and
($v$) nonzero entries of $\x$ have a minimum magnitude, i.e., $\min_{j \in \cJ} |x_j| \geq x_{\mathrm{min}}$ almost surely.
Finally, we define $\kappa_x \triangleq \bbE\lr{|x|}/\sqrt{\bbE \lr{x^2}}$ as a measure of the flatness of $\x$ ($\kappa_x \leq 1$, with $\kappa_x=1$ when all nonzero coefficients are equal~\cite{gribonval2014sparse}).

\textit{Noise distribution:} We make following assumptions on the distribution of noise, which is assumed i.i.d. across data samples:
($i$) $\bbE\lr{\n \n^\top} = \bbE\lr{w^2} \I_m $,
($ii$) $\bbE\lr{\n \x^\top |\cJ}=\bbE\lr{\n \s^\top |\cJ}=\mb{0}$, and
($iii$) magnitude of $\n$ is bounded, i.e., $\|\n\|_2 \leq M_w$ almost surely.

Our goal in this paper is to recover the underlying coordinate dictionaries, $\D^0_k$, from $N$ noisy realizations of tensor data.
To solve this problem, we take the empirical risk minimization approach and define
	\begin{align}
	&f_\y \lrp{\Dks} \triangleq
		 \inf_{\x' \in \bbR^p } \bigg\{
		 \frac{1}{2} \norm{\y - \lrp{\bo \D_k } \x'
		 }_2^2
		 +\lambda\|\x'\|_1 \bigg\},
 		\text{and} \nonum \\
	&F_\Y \lrp{\Dks} \triangleq
		 \frac{1}{N} \sum_{n =1}^N f_{\y_n}\lrp{\Dks} ,
	\end{align}
where $\lambda$ is a regularization parameter. In theory, we can recover the coordinate dictionaries by solving the following regularized optimization program:
	\begin{align}
    \min_{\substack{\D_k \in \mc{D}_k \\ k \in [K]}}
		F_\Y \lrp{\Dks}. \label{eq:f_x}
	\end{align}
More specifically, given desired errors $\lr{\eps_k}_{k=1}^K$, we want a local minimum of~\eqref{eq:f_x} to be attained by coordinate dictionaries $\whD_k \in \mc{B}_{\eps_k}(\D^0_k), k \in [K]$. That is, there exists a set $\{\whD_k\}_{k \in [K]} \subset \lr{\D_k \in \mc{B}_{\eps_k}(\D^0_k)}_{k \in [K]}$ such that $F_\Y(\whD_{1:K}) \leq F_\Y(\Dks)$.\footnote{We focus on the local recovery of coordinate dictionaries (i.e., $\whD_k \in \mc{B}_{\eps_k}(\D^0_k)$) due to ambiguities in the general DL problem. This ambiguity is a result of the fact that dictionaries are invariant to permutation and sign flips of dictionary columns, resulting in equivalent classes of dictionaries. Some works in the literature on conventional overcome this issue by defining distance metrics that capture the distance between these equivalent classes~\cite{agarwal2013exact,agarwal2013learning,arora2013new}.}
To address this problem, we first minimize the statistical risk:
	\begin{align} \label{eq:f_x_asym}
	&\min_{\substack{\D_k \in \mc{D}_k \\ k \in [K]}} f_\bbP \lrp{\Dks} \triangleq
	\min_{\substack{\D_k \in \mc{D}_k \\ k \in [K]}}
		\bbE_\y \lr{ f_\y\lrp{\Dks}}.
	\end{align}
Then, we connect $F_\Y \lrp{\Dks}$ to $f_\bbP \lrp{\Dks}$ using concentration of measure arguments and obtain the number of samples sufficient for local recovery of the coordinate dictionaries. Such a result ensures that any KS-DL algorithm that is guaranteed to converge to a local minimum, and which is initialized close enough to the true KS dictionary, will converge to a solution close to the generating coordinate dictionaries (as opposed to the generating KS dictionary, which is guaranteed by analysis of the vector-valued setup~\cite{gribonval2014sparse}).

\section{Asympototic Identifiability Results}\label{sec:asymp}
In this section, we provide an identifiability result for the KS-DL objective function in \eqref{eq:f_x_asym}. The implications of this theorem are discussed in Section~\ref{sec:discuss}.
\begin{theorem}\label{thm:asymp}
Suppose the observations are generated according to \eqref{eq:obs_model} and the dictionary coefficients follow the separable sparsity model of Section~\ref{sec:model}. Further, assume the following conditions are satisfied:
	\begin{align} \label{eq:cond_k_i_p_i}
	&s_k \leq \frac{p_k}{8\lrp{\norm{\D^0_k}_2+1}^2}, \\\nonum
	&\max_{k \in [K]} \lr{\mu_{s_k}(\D^0_k)} \leq \frac{1}{4} ,  \quad
	\mu_s(\D^0) <\frac{1}{2},
	\end{align}
and
	\begin{align} \label{eq:cond_m_p}
	&\frac{\bbE\lr{x^2}}{M_x \bbE\lr{|x|}} > \frac{24\sqrt{3}(4.5^{K/2})K}{(1-2\mu_s(\D^0))} \nonum\\
	&\qquad \quad \max_{k \in [K]} \lr{
		 \frac{s_k}{p_k}
		 \norm{ {\D^0_k}^\top\D^0_k - \I}_F
		 \lrp{ \norm{\D^0_k}_2+1}}.
	\end{align}
Define
	\begin{align} \label{eq:cond_C_min_C_max}
	&C_{k,\min} \triangleq   8 (3^{\frac{K+1}{2}})\kappa_x^2
		\lrp{\frac{s_k}{p_k}}
		\norm{{\D_k^0}^\top\D_k^0 - \I}_F  \lrp{ \norm{\D^0_k}_2+1} ,
		 \nonum \\
	&C_{\max} \triangleq \frac{1}{3K(1.5)^{K/2}} \frac{\bbE\lr{|x|}}{M_x} (1-2\mu_s(\D^0)).
	\end{align}
Then, the map $\Dks  \mapsto f_{\bbP}\lrp{\Dks}$ admits a local minimum $\wh{\D}=\bo_{k \in [K]} \wh{\D}_k$ such that $\wh{\D}_k \in \mc{B}_{\eps_k}(\D^0_k)$, $k \in [K]$, for any $\eps_k>0$ as long as
	\begin{align} \label{eq:cond_lambda}
	\lambda \leq \frac{x_\mathrm{min}}{8\times 3^{(K-1)/2}},
	\end{align}
	\begin{align} \label{eq:cond_r_i}
	&\frac{\lambda C_{k,\min}}{\bbE\lr{|x|}} < \eps_k < \frac{\lambda C_{\max}}{\bbE\lr{|x|}}, \ k \in [K],
	\end{align}
and
	\begin{align} \label{eq:cond_noise}
	\frac{M_w}{M_x}
		 < 3(1.5)^{K/2} \bigg(\frac{\lambda K C_{\max} }{\bbE\lr{|x|}}
		 - \sum_{k \in [K]} \eps_k\bigg).
	\end{align}
\end{theorem}

\subsection{Discussion}

Theorem~\ref{thm:asymp} captures how the existence of a local minimum for the statistical risk minimization problem depends on various properties of the coordinate dictionaries and demonstrates that there exists a local minimum of $f_{\bbP} \lrp{\Dks}$ that is in local neighborhoods of the coordinate dictionaries. This ensures asymptotic recovery of coordinate dictionaries within some local neighborhood of the true coordinate dictionaries, as opposed to KS dictionary recovery for vectorized observations~\cite[Theorem 1]{gribonval2014sparse}.

We now explicitly compare conditions in Theorem~\ref{thm:asymp} with the corresponding ones for vectorized observations~\cite[Theorem 1]{gribonval2014sparse}.
Given that the coefficients are drawn from the separable sparsity model, the sparsity constraints for the coordinate dictionaries in \eqref{eq:cond_k_i_p_i} translate into
	\begin{align}
	\frac{s}{p} = \prod_{k \in [K]} \frac{ s_k}{ p_k}
		\leq \frac{1}{8^K \prod_k \lrp{\norm{\D^0_k}_2+1}^2} .
	\end{align}
Therefore, we have $\dfrac{s}{p}= \mc{O}\lrp{ \frac{1}{ \prod_k \norm{\D^0_k}_2^2}}=\mc{O}\lrp{\frac{1}{\|\D^0\|_2^2}}$. Using the fact that $\norm{\D^0}_2 \geq \|\D^0\|_F/ \sqrt{m} = \sqrt{p}/\sqrt{m}$, this translates into sparsity order
$	s = \mc{O}\lrp{ m}$. Next, the left hand side of the condition in \eqref{eq:cond_m_p} is less than 1. Moreover, from properties of the Frobenius norm, it is easy to show that
$
	\norm{{\D^0_k}^\top\D_k^0 - \I }_F \geq \sqrt{p_k(p_k-m_k)/m_k}.
$
The fact that $\norm{\D_k^0}_2 \geq \sqrt{p_k}/\sqrt{m_k}$ and the assumption $\mu_{s_k}(\D_k^0)\leq 1/4$ imply that the right hand side of \eqref{eq:cond_m_p} is lower bounded by $\Omega\lrp{ \max_k s_k\sqrt{(p_k-m_k)/m_k^2}}$.
Therefore, Theorem~\ref{thm:asymp} applies to coordinate dictionaries with  dimensions $p_k \leq m_k^2$ and subsequently, KS dictionaries with $p \leq m^2$. Both the sparsity order and dictionary dimensions are in line with the scaling results for vectorized data~\cite{gribonval2014sparse}.

\subsection{Proof Outline}

For given radii $0<\eps_k\leq 2\sqrt{p_k}, k \in [K]$, the spheres $\cS_{\eps_k}(\D^0_k)$ are non-empty. This follows from the construction of dictionary classes, $\cD_k$'s.
Moreover, the mapping $\Dks \mapsto f_{\bbP}\lrp{\Dks}$ is continuous with respect to the Frobenius norm $\FnDk$ on all $\D_k,\D'_k \in \bbR^{m_k\times p_k}, k \in [K]$~\cite{gribonval2015sample}. Hence, it is also continuous on compact constraint sets $\mc{D}_k$'s.
We derive conditions on the coefficients, underlying coordinate dictionaries, $M_w$, regularization parameter, and $\eps_k$'s such that
	\begin{align}  \label{eq:f_p_r_def}
	\Delta f_{\bbP}\lrp{\rks}
		\triangleq \inf_{\D_k \in \cS_{\eps_k}(\D^0_k)}
		\Delta f_{\bbP} \lrp{ \Dks; \Drks } >0.
	\end{align}
This along with the compactness of closed balls $\bar{\mc{B}}_{\eps_k}(\D^0_k)$ and the continuity of the mapping $\Dks \mapsto f_{\bbP}\lrp{\Dks}$ imply the existence of a local minimum of $f_{\bbP}\lrp{\Dks}$ achieved by $\Dhks$ in open balls, $\mc{B}_{\eps_k}(\D_k^0)$'s, $k \in [K]$.

To find conditions that ensure $\Delta f_{\bbP}\lrp{\rks} > 0$, we take the following steps:
given coefficients that follow the separable sparsity model, we can decompose any $\D_\cJ, |\cJ|=s$, as
	\begin{align} \label{eq:Dj_D12}
	\D_\cJ  = \bo \D_{k,\cJ_k},
	\end{align}
where $\ |\cJ_k|=s_k$ for $k \in[K]$.\footnote{The separable sparsity distribution model implies sampling without replacement from columns of $\D_k$.}
Given a generating $\s = \sign(\x)$,
we obtain $\wh{\x}$ by solving $f_\y\lrp{\Dks}$ with respect to $\x'$, conditioned on the fact that $\sign(\wh{\x})=\wh{\s}=\s$. This eliminates the dependency of $f_\y \lrp{\Dks}$ on $\inf_{\x'}$ by finding a closed-form expression for $f_\y\lrp{\Dks}$ given $\wh{\s}=\s$, which we denote as $\phi_\y\lrp{\Dks|\s}$. Defining
\begin{align}
\phi_{\bbP}\lrp{\Dks|\s} \triangleq \bbE\lr{\phi_\y\lrp{\Dks|\s}},
\end{align}
we expand $ \Delta \phi_{\bbP}\lrp{ \Dks;\Drks | \s}$  using \eqref{eq:Dj_D12} and separate the terms that depend on each radius $\eps_k = \FrDk$ to obtain conditions for sparsity levels $s_k, k\in [K]$, and coordinate dictionaries such that $\Delta \phi_{\bbP} \lrp{ \Dks;\Drks | \s} >0$.
Finally, we derive conditions on $M_w$, coordinate dictionary coherences and $\eps_k$'s that ensure $\wh{\s}=\s$ and $\Delta f_{\bbP} \lrp{ \Dks;\Drks } = \Delta \phi_{\bbP} \lrp{\Dks;\Drks | \s}$.

\begin{remark}
The key assumption in the proof of Theorem~\ref{thm:asymp} is expanding $\D_{\cJ}$ according to~\eqref{eq:Dj_D12}. This is a consequence of the separable sparsity model for dictionary coefficients. For a detailed discussion on the differences between the separable sparsity model and the random sparsity model for tensors, we refer the readers to our earlier work~\cite{shakeri2016minimax}.
\end{remark}

\begin{remark}
Although some of the forthcoming lemmas needed of Theorem~\ref{thm:asymp} impose conditions on $\D_k$'s as well as true coordinate dictionaries $\D^0_k$'s, we later translate these conditions exclusively in terms of $\D_k^0$'s and $\eps_k$'s.
\end{remark}

The proof of Theorem~\ref{thm:asymp} relies on the following propositions and lemmas. The proofs of these are provided in Appendix A.

\begin{proposition} \label{prop:1}
Suppose the following inequalities hold for $k \in [K]$:
	\begin{align} \label{eq:cond_delt_s_prop}
	s_k \leq \frac{p_k}{8(\|\D_k^0\|_2+1)^2}\quad \text{and} \quad
	\max_{k\in [K]}&\lr{\delta_{s_k}(\D_k^0)}\leq \frac{1}{4} .
	\end{align}
Then, for
	\begin{align} \label{eq:cond_lamb_1}
	\bar{\lambda} \triangleq \dfrac{\lambda}{\bbE\lr{|x|}}\leq \dfrac{1}{8\times 3^{(K-1)/2}},
	\end{align}
any collection of $\lr{ \eps_k: \eps_k \leq 0.15, k \in [K]}$, and for all $\D_k \in \cS_{\eps_k}(\D_k^0)$, we have :
	\begin{align} \label{eq:delt_p_LB}
	&\Delta \phi_{\bbP} \lrp{ \Dks;\Drks | \s}
		\geq  \frac{s\bbE\{x^2\}}{8} \sum_{k \in [K]}
		\frac{\eps_k}{p_k}  \lrp{\eps_k- \eps_{k,\min}(\bar{\lambda})},
	\end{align}
where
	\begin{align*}
	&\eps_{k,\min} (\bar{\lambda})
	\triangleq  \frac{3^{(K-1)/2}}{2} \lrp{1.5^{\frac{K-1}{2}}
		+2^{(K+1)}\bar{\lambda} } \bar{\lambda} C_{k,\min}.
	\end{align*}
In addition, if
	\begin{align}  \label{eq:cond_lamb_2}
	\bar{\lambda} \leq \frac{0.15}{\max_{k \in [K]} C_{k,\min}},
	\end{align}
then $\eps_{k,\min}(\bar{\lambda})<0.15$.
Thus, $\Delta \phi_\bbP\lrp{ \Dks;\Drks | \s}  > 0 $ for all $\eps_k \in (\eps_{k,\min}(\bar{\lambda}),0.15], k \in [K]$.
\end{proposition}

The proof of Proposition \ref{prop:1} relies on the following lemmas as well as supporting lemmas from the analysis of vectorized data~\cite[Lemmas~4,6,7,15,16]{gribonval2014sparse}.

\begin{lemma} \label{def:P_H_PS}
Let $\D = \boDk$ where $\delta_s(\D_k)<1$ for $k \in [K]$, and $\cJ$ be a support set generated by the separable sparsity model. Then any $\D_\cJ, |\cJ|=s$, can be decomposed as
$\D_\cJ  = \bo \D_{k,\cJ_k}$,
where $\ |\cJ_k|=s_k$ and $\rnk (\D_{k,\cJ_k})=s_k$, for $k \in[K]$. Also, the following relations hold for this model:\footnote{The equations follow from basic properties of the Kronecker product~\cite{horn2012matrix}.}
	\begin{align} \label{eq:P_H_Ps}
	\PD = \bo \PDk,   \PsD = \bo \PsDk,  \HD = \bo \HDk,
	\end{align}
where $\bP$ and $\bH$ are defined in Section~\ref{subsec:notation}.
\end{lemma}

\begin{lemma} \label{lem:Dtld}
Given $\Dks$ and $\Drks$, the difference
	\begin{align} \label{eq:otD_otDp}
	&\boDk - \boDrk \nonum\\
	&\qquad = \sum _{k \in [K]} \wt{\D}_{k,1} \otimes \dots \otimes
		\lrp{\D_k - \D^0_k}	\otimes \dots \otimes \wt{\D}_{k,K},
	\end{align}	
where without loss of generality, each $\wt{\D}_{k,i}$ is equal to either $\D^0_i$ or $\D_i$, for $k \in [K]$.
\end{lemma}

We drop the $k$ index from $\wD_{k,i}$ for ease of notation throughout the rest of the paper.

\begin{lemma} \label{lemma:f_closedform}
Let $\s \in \{-1,0,1\}^p$ be an arbitrary sign vector and $\cJ = \cJ(\s)$ be its support. Define\footnote{The quantity $\phi_\y\lrp{\Dks|\s}$ is not equal to $\phi_\y\lrp{\Dks}$ conditioned on $\s$ and the expression is only used for notation.}
	\begin{align} \label{eq:delt_inf}
	\phi_\y\lrp{\Dks|\s} \triangleq \inf_{\substack{\x \in \bbR^p \\ \supp(\x) \subset \cJ}}
		\frac{1}{2} \norm{\y - \lrp{\boDk} \x }_2^2+ \lambda {\s}^\top\x.
	\end{align}
If $\D_{k,\cJ_k}^\top \D_{k,\cJ_k}$ is invertible for $k \in [K]$, then $\wh{\x}$ minimizes $\phi_\y\lrp{\Dks|\s} $, where
	\begin{align}  \label{eq:xhat_closedform}
	\wh{\x}_\cJ = \lrp{\bo \PsDk } \y  - \lambda \lrp{ \bo \big( \D_{k,\cJ_k}^\top\D_{k,\cJ_k} \big)^{-1} }\s_\cJ,
	\end{align}
and $\wh{\x}_{\cJ^c} = \mb{0}$. Thus, $\phi_\y\lrp{\Dks |\s}$ can be expressed in closed form as:
	\begin{align} \label{eq:delt_closedform}
	&\phi_\y\lrp{\Dks|\s} = \frac{1}{2}\|\y\|_2^2
		- \frac{1}{2} \y^\top
		\lrp{ \bo \PDk}\y  \nonum\\
	&\ + \lambda {\s}_\cJ^\top
		\lrp{ \bo \PsDk  } \y
		-\frac{\lambda^2}{2}{\s}_\cJ^\top \lrp{ \bo \HDk} \s_\cJ.
	\end{align}
\end{lemma}

\begin{lemma} \label{lem:exp_phi}
Assume $\max\lr{ \delta_{s_k}(\D_k^0),\delta_{s_k}(\D_k)}<1$ for $k \in [K]$ and let $\wD_k$ be equal to either $\D^0_k$ or $\D_k$.
For
	\begin{align}
	\Delta \phi_{\bbP} \lrp{ \Dks;\Drks \big| \s}
		\triangleq \phi_{\bbP} \lrp{ \Dks| \s}
		- \phi_{\bbP} \lrp{ \Drks| \s},
	\end{align}
we have
	\begin{align}  \label{eq:delt_ph_2}
	&\Delta \phi_{\bbP} \lrp{ \Dks;\Drks \big| \s} \nonum\\
	& =  \frac{\bbE\{x^2\}}{2}
		\sum_{k \in [K]} \bbE_{\cJ_1} \lr{ \tr \lrb{ {\D^0_1}^\top \bP_{\wt{\D}_{1,\cJ_1}}\D^0_1} }\dots  \nonum\\
	&\qquad \qquad \qquad \qquad \bbE_{\cJ_k} \lr{ \tr \lrb{  {\D^0_k}^\top(\I_{m_k} - \PDk)\D^0_k}} \nonum \\
	&\qquad \qquad \qquad \qquad \dots
		\bbE_{\cJ_K} \lr{ \tr \lrb{{\D^0_K}^\top \bP_{\wt{\D}_{K,\cJ_K}}\D^0_K}}
		\nonum \\
	&- \lambda \bbE\{|x|\}
		\sum_{k \in [K]} \bbE_{\cJ_1} \lr{ \tr \lrb{ {\wt{\D}_{1,\cJ_1}}^+\D^0_1}} \dots  \nonum\\
	&\quad \bbE_{\cJ_k} \lr{ \tr \lrb{\I_{s_k} - \PsDk \D^0_k}} \dots
		\bbE_{\cJ_K} \lr{ \tr \lrb{  {\wt{\D}_{K,\cJ_K}}^+ \D^0_K}} \nonum \\
	&+  \frac{\lambda^2}{2}
		\sum_{k \in [K]} \bbE_{\cJ_1} \lr{ \tr \lrb{ \bH_{\wt{\D}_1,\cJ_1}} }\dots  \nonum\\
	&\quad \bbE_{\cJ_k} \lr{ \tr \lrb{ \bH_{\D^0_{k,\cJ_k}}  - \HDk }	} \dots
		\bbE_{\cJ_K} \lr{ \tr \lrb{ \bH_{\wt{\D}_{K,\cJ_K}}}}.
	\end{align}
\end{lemma}

\begin{lemma} \label{lem:UB_mu_delt_RIP}
For any $\D_k \in \cD_k$ satisfying $\RIP$ of order $s_k$, given $\cJ_k \subset [p_k]$ and $ |\cJ_k|=s_k$, the following relations hold:
	\begin{align}
	\norm{\D_{k,\cJ_k}}_2 &= \norm{{\D_{k,\cJ_k}}^\top}_2
		\leq \sqrt{1+\delta_{s_k}(\D_k)}, \label{eq:l2_delt} \\
	\delta_{s_k}(\D_k) &\leq \mu_{s_k-1}(\D_k). \label{eq:delt_mu}
	\end{align}
\end{lemma}

\begin{lemma}[Lemma 4~\cite{gribonval2014sparse}]
\label{lem:H_Dps}
Let $\D_k$'s be coordinate dictionaries such that $\delta_{s_k}(\D_k)<1$. Then for any $\cJ_k \subset p_k, |\cJ_k|=s_k$, $\HDk$ exists and
	\begin{align} \label{eq:pso_cond}
	&\norm{\HDk}_2 \leq \frac{1}{1-\delta_{s_k}(\D_k)}, \quad
	\norm{\PsDk}_2 \leq \frac{1}{\sqrt{1-\delta_{s_k}(\D_k)}},
	\end{align}
and for any $\D_k'$ such that $\FnDk \leq \eps_k < \sqrt{1-\delta_{s_k}(\D_k)}$:
		\begin{align} \label{eq:cond_delt_r_i}
		&1-\delta_{s_k}(\D'_k) \geq (\sqrt{1-\delta_{s_k}(\D_k)} - \eps_k)^2 \triangleq 1-\delta_k.
		\end{align}
\end{lemma}

\begin{lemma}[Lemma 6~\cite{gribonval2014sparse}]
\label{lem:D_12_tet}
Given any $\D_k^1,\D_k^2 \in \cD_k$, there exist $\mb{V}_k \in \bbR^{m_k \times p_k}$ with $\diag{{\D^1_k}^\top \mb{V}_k}=\mb{0}$ and $\diag{\mb{V}_k^\top \mb{V}_k}=\I_{p_k}$ and a vector $\tet_k \triangleq \tet_k(\D_k^1,\D_k^2) \in [0,\pi]^{p_k}$, such that
	\begin{align}
	\D_k^2 = \D_k^1 \mb{C}_k (\tet_k) + \mb{V}_k \mb{S}_k(\tet_k),
	\end{align}
where $\mb{C}_k (\tet_k) \triangleq \Diag{\cos(\tet_k) }$ and $\mb{S}_k (\tet_k)  \triangleq \Diag{\sin(\tet_k) }$. Moreover,
	\begin{align} \label{eq:tet_rk}
	&\frac{2}{\pi}\theta_{k,j} \leq \|\bd^2_{k,j} - \bd^1_{k,j} \|_2
		= 2\sin \lrp{\frac{\theta_{k,j}}{2}} \leq\theta_{k,j}, \text{and} \nonum \\
	&\frac{2}{\pi} \|\tet_k\|_2 \leq \|\D_k^2 - \D_k^1 \|_F \leq \|\tet_k\|_2 ,
	\end{align}
where $j \in [p_k]$.
Similarly, there exists $\mb{V}_k'$ such that $\D_k^1 = \D_k^2 \mb{C}_k (\tet_k) + \mb{V}'_k \mb{S}_k(\tet_k)$, where $\diag{{\D^2_k}^\top \mb{V}'_k}=\mb{0}$.
\end{lemma}

\begin{lemma} \label{lem:A_B_delt}
Fix $\Dks$ and $\Drks$, and suppose $\lr{A_k},\lr{B_k},\lr{\delta_k}$ satisfy the following:
	\begin{align} \label{eq:A_B_Delk}
	&A_k \geq \max\lr{ \|\D_k^\top \D_k - \I_{p_k}\|_F,\|{\D_k^0}^\top \D_k^0 - \I_{p_k}\|_F } , \nonum \\
	&B_k \geq \max\lr{ \|\D_k\|_2, \|\D_k^0\|_2 }, \text{and}\nonum \\
	&\delta_k \geq \max\lr{ \delta_{s_k}(\D_k),\delta_{s_k}(\D_k^0) }.
	\end{align}
Then for all $ \tet_k \triangleq \tet_k(\D_k,\D_k^0), k \in [K]$, we have
	\begin{align}  \label{eq: delt_ph_3}
	&\Delta\phi_{\bbP}\lrp{ \Dks;\Drks|\s}  \nonum\\
	&\geq \frac{s\bbE\{x^2\}}{2}
	\sum_{k \in [K]} \frac{\|\tet_k\|_2 }{p_k}
		\bigg[\|\tet_k\|_2
		\bigg( 1 - \frac{s_k}{p_k} \frac{B^2_k}{1-\delta_k}
		-\bar{\lambda} \kappa_x^2
		\delta_{-k}\bigg)
		\nonum \\
	&\qquad \quad -\bigg(\delta_{-k}
		+2\bar{\lambda}\prod_{i \in [K]} \frac{1}{1-\delta_i}\bigg)
		\bar{\lambda} \kappa_x^2
		\frac{s_k}{p_k} \frac{2A_kB_k}{1-\delta_k}\bigg],
	\end{align}
where $\bar{\lambda} \triangleq \dfrac{\lambda}{\bbE\lr{|x|}} $ and $\delta_{-k} \triangleq \prod_{\substack{i \in [K]\\ i \neq k}}
		\sqrt{\dfrac{1+\delta_i}{1-\delta_i}} $.
\end{lemma}

Proposition~\ref{prop:1} shows $\Delta\phi_{\bbP}\lrp{ \Dks;\Drks|\s} > 0 $. However, given $\wh{\x}$, the solution of $\phi_\y\lrp{\Dks|\s}$, $\wh{\s} = \sign\lrp{\wh{\x}}$ is not necessarily equal to the sign of the generating $\s$. We derive conditions that ensure $\wh{\x}$ is almost surely the unique minimizer of $f_\y\lrp{\Dks}$ and $\wh{\s}=\s$.
We introduce the following proposition for this purpose.

\begin{proposition} \label{prop:3}
Let the generating coordinate dictionaries $\{ \D_k^0 \in\cD_k\}$ satisfy:
	\begin{align} \label{eq:delt_cond}
	\mu_{s}(\D^0) <  \frac{1}{2} , \quad  \max_k\{ \delta_{s_k}(\D_k^0)\} <  \frac{1}{4} .
	\end{align}
Suppose $\bar{\lambda} = \dfrac{\lambda}{\bbE\lr{|x|}}\leq \dfrac{x_{\min}}{2\bbE\lr{|x|}}$ and
	\begin{align} \label{eq:prop_r_max}
	\max_{k\in [K]}\{\eps_k\} \leq \min\lr{ \bar{\lambda}C_{\max}, 0.15}.
	\end{align}
If the following is satisfied:
	\begin{align} \label{eq:M_eps_M_al}
	\frac{M_w}{M_x}
		 < 3(1.5)^{K/2} \bigg( \bar{\lambda} K C_{\max}  - \sum_{k \in [K]} \eps_k\bigg),
	\end{align}
then for any $\Dks$ such that $\D_k \in \cS_{\eps_k}(\D^0_k)$, for $k \in [K]$, $\wh{\x}$ that is defined in~\eqref{eq:xhat_closedform} is almost surely the minimizer of the map $\x' \mapsto \frac{1}{2}\norm{\y- \lrp{\boDk} \x'}_2^2 +\lambda \|\x'\|_1 $ and 	
$\Delta\phi_{\bbP}\lrp{\Dks;\Drks|\s} = \Delta f_{\bbP} \lrp{\Dks;\Drks}$.
\end{proposition}

\begin{remark}
Note that $\mu_s(\D^0) < \frac{1}{2}$ in \eqref{eq:delt_cond} can be satisfied by ensuring that the right hand side of~\eqref{eq:mu_s} is less than $\frac{1}{2}$. One way this can be ensured is by enforcing strict conditions on coordinate dictionaries; for instance, $\mu_{s_k}(\D^0_k)\leq \frac{1}{2^K}$.
\end{remark}

The proof of Proposition~\ref{prop:3} relies on the following lemmas and~\cite[Lemmas 10--13]{gribonval2014sparse}.

\begin{lemma}[Lemma 13~\cite{gribonval2014sparse}]
\label{lem:a_hat_min_cond}
Assume $\mu_s(\D) <\dfrac{1}{2}$. If
	\begin{align} \label{eq:lem8_cod}
	\min_{j \in \cJ} \left| x_j \right| \geq 2\lambda, \
		\text{and} \
		\norm{\y - \D \x}_2 < \lambda (1-2\mu_s(\D))
	\end{align}
hold for generating $\x$, then $\wh{\x}$ defined in \eqref{eq:xhat_closedform} is the unique solution of $\min_{\x'} \frac{1}{2}\norm{\y - \lrp{\boDk} \x' }_2  +\lambda\|\x'\|_1$.
\end{lemma}

\begin{lemma} \label{lem:mu_mu0_rel}
For any $\D^0=\boDrk$ and $\D = \boDk$  such that
$\D_k \in \bar{\mc{B}}_{\eps_k}(\D^0_k)$, for $k \in [K]$, suppose the following inequalities are satisfied:
	\begin{align} \label{eq:UB_mu_delt}
	\max_{k \in [K]} \{\delta_{s_k}(\D_k^0)\} \leq \frac{1}{4},
		\quad \text{and}
		\quad	\max_{k \in [K]} \eps_k \leq 0.15.
	\end{align}
Then, we have
	\begin{align} \label{eq:mu_mu0}
	\mu_s(\D) \leq  \mu_s(\D^0) + 2(1.5)^{K/2}\sqrt{s} \bigg(\sum_{k \in [K]} \eps_k\bigg).
	\end{align}	
\end{lemma}

\begin{IEEEproof}[Proof of Theorem~\ref{thm:asymp}]
To prove this theorem, we use Proposition~\ref{prop:1} to show that $\Delta \phi_{\bbP}\lrp{ \Dks;\Drks | \s} > 0$, and then use Proposition~\ref{prop:3} to show that $\Delta \phi_{\bbP}\lrp{ \Dks;\Drks | \s} =\Delta f_{\bbP}\lrp{\Dks;\Drks}$.
The assumptions in \eqref{eq:cond_k_i_p_i} ensure that the conditions in \eqref{eq:cond_delt_s_prop} and \eqref{eq:delt_cond} are satisfied for Proposition~\ref{prop:1} and Proposition~\ref{prop:3}, respectively. Assumptions \eqref{eq:cond_m_p} and \eqref{eq:cond_lambda} ensure that the conditions in \eqref{eq:cond_lamb_1} and \eqref{eq:cond_lamb_2}  are satisfied for Proposition~\ref{prop:1},  $\bar{\lambda}\leq \dfrac{x_{\min}}{2\bbE\lr{|x|}}$ holds for Proposition~\ref{prop:3}, and $\max_{k \in [K]}\{C_{k,\mathrm{min}}\}<C_{\max}$. Hence, according to Proposition~\ref{prop:1},  $\Delta \phi_{\bbP}\lrp{ \Dks;\Drks | \s} > 0 $ for all $\eps_k \in (\bar{\lambda}C_{k,\min},0.15], k \in [K]$. Finally, using the assumption in \eqref{eq:cond_noise} implies $\Delta \phi_{\bbP}\lrp{ \Dks;\Drks | \s} =\Delta f_{\bbP}\lrp{\Dks;\Drks}$ for all $\eps_k \leq \bar{\lambda}C_{\max}, k \in [K]$. Furthermore, the assumption in \eqref{eq:cond_lambda} implies $C_{\max}\bar{\lambda} \leq 0.15$. Consequently, for any $\lr{ \eps_k>0,k \in [K]}$ satisfying the conditions in \eqref{eq:cond_r_i}, $\Dks \rightarrow f_{\bbP}\lrp{\Dks}$ admits a local minimum $\wh{\D}= \bo \wh{\D}_k$ such that $\wh{\D}_k \in \mc{B}_{\eps_k}(\D^0_k), k \in [K]$.
\end{IEEEproof}

\section{Finite Sample Identifiability Results} \label{sec:finite}

We now focus on leveraging Theorem~\ref{thm:asymp} and solving~\eqref{eq:f_x} to derive finite-sample bounds for KS dictionary identifiability. Compared to Gribonval et al.~\cite{gribonval2014sparse}, who use Lipschitz continuity of the objective function with respect to the larger KS dictionary, our analysis is based on  ``coordinate-wise Lipschitz continuity" with respect to the coordinate dictionaries.

\begin{theorem}\label{thm:finite_n}
Suppose the observations are generated according to \eqref{eq:obs_model} and the dictionary coefficients follow the separable sparsity model of Section~\ref{sec:model} such that \eqref{eq:cond_k_i_p_i} to \eqref{eq:cond_noise} are satisfied. Next, fix any $\xi \in (0,\infty)$. Then, for any number of observations satisfying
	\begin{align} \label{eq:smplCmp}
	N = \max_{k \in [K]}
		&\Omega \bigg(
		\frac{p_k^2 (	\xi+m_kp_k) } {(\eps_k - \eps_{k,\min}(\bar{\lambda}))^2}
		\nonum\\
	&\qquad \bigg( \frac{2^{K}(1 + \bar{\lambda}^2) M_x^2}
		{s^2\bbE\{x^2\}^2}
		+ \bigg(\frac{M_w}{s\bbE\{x^2\}} \bigg)^2 \bigg) \bigg),
	\end{align}
with probability at least $1-e^{-	\xi}$, $\Dks \mapsto F_\Y\lrp{\Dks}$ admits a local minimum $\wh{\D}=\bo \wh{\D}_k$ such that $\wh{\D}_k \in \mc{B}_{\eps_k}(\D^0_k)$, for $k \in [K]$.
\end{theorem}

\subsection{Discussion}
Let us make some remarks about implications of Theorem~\ref{thm:finite_n}. First, sample complexity has an inverse relationship with signal to noise ratio ($\SNR$),\footnote{Sufficient conditioning on $N$ implies $\mc{O}$-scaling for sample complexity.} which we define as
	\begin{align}
	\SNR \triangleq \frac{\bbE\{\|\x\|_2^2\}}{\bbE\{\|\n\|^2_2\}} = \frac{s\bbE\{x^2\}}{m\bbE\{w^2\}}.
	\end{align}
Looking at the terms on the right hand side of~\eqref{eq:smplCmp} in Theorem~\ref{thm:finite_n}, $M_x/(s\bbE\lr{x^2})$ is related to the deviation of $\|\x\|_2$ from its mean, $\bbE\lr{\|\x\|_2}$, and depends on the coefficient distribution, while $M_w/(s\bbE\lr{x^2})$ is related to $1/\SNR$ and depends on the noise and coefficient distributions.

Second, we notice dependency of sample complexity on the recovery error of coordinate dictionaries. We can interpret $\eps_k$ as the recovery error for $\D^0_k$. Then, the sample complexity scaling in \eqref{eq:smplCmp} is proportional to $\max_k \eps_k^{-2}$.
We note that the sample complexity results obtained in~\cite{gribonval2014sparse} that are independent of $\eps \triangleq \norm{\D-\D^0}_F$ only hold for the noiseless setting and the dependency on $\eps^{-2}$ is inevitable for noisy observations~\cite{gribonval2014sparse}. Furthermore, given the condition on the range of $\eps_k$'s in \eqref{eq:cond_r_i}, $\eps_k$'s cannot be arbitrarily small, and will not cause $N$ to grow arbitrarily large.

Third, we observe a linear dependence between the sample complexity scaling in \eqref{eq:smplCmp} and coordinate dictionaries' dimensions, i.e., $\max_k \mc{O}(m_k p_k^3)$. Comparing this to the $\mc{O}(mp^3)=\mc{O}\lrp{\prod_k m_kp_k^3}$ scaling in the unstructured DL problem~\cite{gribonval2014sparse}, the sample complexity in the KS-DL problem scales with the dimensions of the largest coordinate dictionary, as opposed to the dimensions of the larger KS dictionary.

\begin{table}
\caption{\small Comparison of upper and lower bounds on the sample complexity of dictionary learning for vectorized DL and KS DL.}
\label{table:1}
\centering
\begin{tabular}{l|C{1.7cm}|c| N} \cline{2-3}
 & Vectorized DL & KS DL & \\ [20pt] \hline
	\multicolumn{1}{|c|} {Minimax Lower Bound}
	&$\dfrac{mp^2}{\eps^2}$~\cite{jung2015minimax}
	&$\dfrac{p\sum_{k } m_k p_k}{\eps^2 }$~\cite{shakeri2016arxiv}
	& \\ [20pt]\hline
	\multicolumn{1}{|c|} {Achievability Bound} & $\dfrac{mp^3}{\eps^2}$~\cite{gribonval2014sparse}
	& $\max\limits_k \dfrac{m_kp_k^3}{\eps_k^2} $ & \\ [20pt]\hline
\end{tabular}
\end{table}

We also compare this sample complexity upper bound scaling to the sample complexity lower bound scaling in our previous work~\cite[Corollary 1]{shakeri2016minimax}, where we obtained $N = \Omega\lrp{p\sum_k m_kp_k\eps^{-2}/K}$ as a \emph{necessary condition for recovery of KS dictionaries}.\footnote{We have the following relation between $\eps$ and $\eps_k$'s:
	\begin{align*}
	\eps 	\leq	\sum_{k \in [K]} \bigg(
		\prod_{\substack{i \in [K]\\i \neq k}} \norm{\wD_k}_F \bigg)
		\norm{\D_k-\D^0_k}_F
		\leq \sqrt{p} \sum_{k \in [K]} \eps_k.
	\end{align*}
Assuming all $\eps_k$'s are equal, this then implies $\eps_k^2 \geq \eps^2/(K^2 p)$.}
In terms of overall error $\eps$, our result translates into $N = \max_k \Omega\lr{ 2^K K^2p(m_kp_k^3)\eps^{-2}}$ as a \emph{sufficient} condition for recovery of coordinate dictionaries. The lower bound depended on the average dimension of the coordinate dictionaries, $\sum_k m_kp_k/K$, whereas we observe here a dependence on the dimensions of the coordinate dictionaries in terms of the maximum dimension, $\max_k m_kp_k$. We also observe an increase of order $\max_k p_k^2$ in the sample complexity upper bound scaling. This gap suggests that tighter bounds can be obtained for lower and/or upper bounds. A summary of these results is provided in Table~\ref{table:1} for a fixed $K$.

\subsection{Proof Outline}

We follow a similar approach used in~\cite[Theorem 2]{gribonval2014sparse} for vectorized data.
We show that, with high probability,
	\begin{align}
	\Delta F_\Y (\rks) \triangleq \inf_{\D_k \in \cS_{\eps_k}(\D_k^0)} \Delta F_\Y \lrp{ \Dks;\Drks}
	\end{align}
converges uniformly to its expectation,
	\begin{align}
	\Delta f_{\bbP}(\rks)\triangleq \inf_{\D_k \in \cS_{\eps_k}(\D^0_k)}
		\Delta f_{\bbP} \lrp{ \Dks; \Drks }.
	\end{align}
In other words, with high probability,	
	\begin{align}
	\lra{\Delta F_\Y (\rks) - \Delta f_{\bbP}(\rks) } \leq \eta_N,
	\end{align}
where $\eta_N$ is a parameter that depends on the probability and other parameters in the problem.
This implies $\Delta F_\Y (\rks) \geq \Delta f_{\bbP}(\rks) - 2\eta_N $.
In Theorem~\ref{thm:asymp}, we obtained conditions that ensure $\Delta f_{\bbP}(\rks)> 0$. Thus, if $2\eta_N < \Delta f_{\bbP}(\rks)$ is satisfied, this implies $\Delta F_\Y (\rks)> 0$, and we can use arguments similar to the proof of Theorem~\ref{thm:asymp} to show that
$\Dks  \mapsto F_\Y\lrp{\Dks}$ admits a local minimum $\wh{\D}=\bo \wh{\D}_k$, such that $\wh{\D}_k \in \mb{B}_{\eps_k}(\D^0_k)$, for $k \in [K]$.

In Theorem~\ref{thm:asymp}, we showed that under certain conditions, $f_{\bbP}(\Dks;\Drks) = \Delta \phi_\bbP\lrp{\Dks;\Drks|\s}$.
To find $\eta_N$, we uniformly bound deviations of $\Dks \mapsto \Delta \phi_\y\lrp{\Dks;\Drks|\s}$ from its expectation on $\lr{ \cS_{\eps_k}(\D^0_k)}_{k=1}^K$.
Our analysis is based on the \textit{coordinate-wise Lipschitz continuity} property of $\Delta \phi_\y\lrp{\Dks;\Drks|\s}$ with respect to coordinate dictionaries. Then, to ensure $ 2\eta_N < \Delta \phi_\bbP\lrp{\Dks;\Drks|\s}$, we show that $2\eta_N$ is less than the right-hand side of~\eqref{eq:delt_p_LB} and obtain conditions on the sufficient number of samples based on each coordinate dictionary dimension and recovery error.

The proof of Theorem~\ref{thm:finite_n} relies on the following definition and lemmas. The proofs of these are provided in Appendix B.

\begin{definition}[Coordinate-wise Lipschitz continuity]
A function $f: \cD_1 \times \dots \times \cD_K \rightarrow \bbR$ is coordinate-wise Lipschitz continuous with constants $(L_1,\dots,L_K)$ if there exist real constants $\lr{L_k \geq 0}_{k=1}^K$, such that for $\lr{\D_k,\D'_k \in \cD_k}_{k=1}^K$:
	\begin{align}
	\lra{ f\lrp{\Dks} - f\lrp{\Dkp} }
		\leq \sum_{k \in [K]} L_k \norm{\D_k - \D'_k}_F.
	\end{align}
\end{definition}

\begin{lemma}[Rademacher averages~\cite{gribonval2014sparse}]\label{lem:rad}
Consider $\mc{F}$ to be a set of measurable functions on measurable set $\cX$ and $N$ i.i.d. random variables $X_1,\dots,X_N \in \cX$. Fix any $\xi \in (0,\infty)$. Assuming all functions are bounded by $B$, i.e., $|f(X)|\leq B$, almost surely, with probability at least $1-e^{-\xi}$:
	\begin{align} \label{eq:rad_gau}
	& \sup_{f \in \mc{F}} \bigg( \frac{1}{N}
		\sum_{n \in [N] } f \lrp{X_n}
		- \bbE_{X} \lr{ f \lrp{X}} \bigg)  \nonum \\
	&\quad
	\leq 2\sqrt{\frac{\pi}{2}} \bbE_{X,\beta_{1:N}}
		\bigg\{ \sup_{f \in \mc{F}}
		\bigg( \frac{1}{N}
		\sum_{n \in [N] } \beta_n f \lrp{ X_n} \bigg) \bigg\}
		+ B\sqrt{\frac{2\xi}{N}},
	\end{align}
where $\beta_{1:N}$'s are independent standard Gaussian random variables.
\end{lemma}

\begin{lemma} \label{lemma:delt_m_T_dev}
Let $\mc{H}$ be a set of real-valued functions on $\D_k \in \overline{\mc{B}}_{\eps_k}(\D^0_k), k \in [K]$, that are bounded by $B$ almost everywhere and are coordinate-wise Lipschitz continuous with constants $(L_1,\dots,L_K)$ .
Let $h_1,h_2,\dots,h_N$ be independent realizations from $\mc{H}$ with uniform Haar measure on $\mc{H}$. Then, fixing $\xi \in (0,\infty)$, we have with probability greater than $1-e^{-\xi}$ that:
	\begin{align} \label{eq:dev}
	&\sup_{\substack{ \D_k \in \overline{\mc{B}}_{\eps_k}(\D^0_k) \\ k \in [K]}}
		 \bigg| \frac{1}{N} \sum_{n \in [N]}
		 h_n(\Dks) - \bbE \lr{ h(\Dks)} \bigg|   \nonum\\
	&\qquad \quad \leq 4\sqrt{\frac{\pi}{2N}} \bigg(\sum_{k \in [K]} L_k\eps_k \sqrt{Km_kp_k} \bigg)
		+  B \sqrt{\frac{2\xi}{N}}.
	\end{align}
\end{lemma}

\begin{lemma}[Lemma 5~\cite{gribonval2014sparse}]
\label{lem:H_Ps}
For any $\delta_k<1$, $\D_k,\D_k'$ such that $\max(\delta_{s_k}(\D_k),\delta_{s_k}(\D'_k))\leq \delta_k$, and $\cJ_k \subset p_k, |\cJ_k|=s_k$, we have
	\begin{align} \label{eq:PH_PHp}
	&\|\I - \PsDk \D'_{k,\cJ_k}\|_2 \leq (1-\delta_k)^{-1/2} \FnDk, \nonum \\
	&\|\HDk-\HDkp \|_2 \leq 2(1-\delta_k)^{-3/2} \FnDk, \nonum \\
	&\|\PsDk - \PsDkp \|_2\leq 2(1-\delta_k)^{-1} \FnDk,\text{and}\nonum \\
	&\|\PDk -\PDkp \|_2 \leq 2(1-\delta_k)^{-1/2} \FnDk.
	\end{align}
\end{lemma}

\begin{lemma} \label{lem:phi_m_T1_lip}
Consider $\D^0_k \in \cD_k$ and $\eps_k$'s such that $\eps_k < \sqrt{1-\delta_{s_k}(\D_k^0)}$, for $k \in [K]$ and define
 $\sqrt{1-\delta_k} \triangleq \sqrt{1-\delta_{s_k}(\D_k^0)} - \eps_k>0$. The function $\Delta \phi_\y\lrp{\Dks;\Drks|\s}$ is almost surely coordinate-wise Lipschitz continuous on $\lr{ \mc{B}_{\eps_k}(\D^0_k)}_{k=1}^K$ with Lipschitz constants
 	\begin{align} \label{eq:lipsch_const_h}
 	L_k \triangleq (1-\delta_k)^{-1/2}
		 \bigg(& M_x \bigg( \prod_{k \in [K]} \sqrt{1+\delta_{s_k}(\D^0_k)}\bigg)
		+M_w \nonum\\
	&+ \lambda\sqrt{s}
		 \prod_{k \in [K] } (1-\delta_k)^{-1/2}  \bigg)^2 ,
 	\end{align}
and $\lra{\Delta \phi_\y\lrp{\Dks;\Drks|\s}}$ is almost surely bounded on $\lr{ \mc{B}_{\eps_k}(\D^0_k)}_{k=1}^K$ by $\sum_{k \in [K]} L_k\eps_k$.
\end{lemma}

\begin{IEEEproof}[Proof of Theorem 2]
From Lemmas~\ref{lemma:delt_m_T_dev} and \ref{lem:phi_m_T1_lip}, we have that with probability at least $1-e^{-\xi}$:
	\begin{align} \label{eq:delt_finite_UB}
	& \sup_{\substack{\D_k \in \overline{\mc{B}}_{\eps_k}(\D^0_k)\\ k \in [K]}}
		\big| \Delta \phi_\y\lrp{\Dks;\Drks|\s} - \Delta \phi_\bbP \lrp{\Dks;\Drks|\s} \big|  \nonum\\
	&\qquad \quad \quad \leq \sqrt{\frac{2}{N}}\sum_{k \in [K]} L_k\eps_k \lrp{ 2\sqrt{\pi m_kp_k}
		+ \sqrt{\xi}},
	\end{align}
where $L_k$ is defined in \eqref{eq:lipsch_const_h}.
From \eqref{eq:delt_finite_UB}, we obtain $ \Delta \phi_\y\lrp{\Dks;\Drks|\s} >\Delta \phi_\bbP \lrp{\Dks;\Drks|\s} - 2\eta_N$ where $\eta_N = \sqrt{\frac{2}{N}}\sum_{k \in [K]} L_k\eps_k \lrp{ 2\sqrt{\pi m_kp_k}
		+ \sqrt{\xi}}$. In Theorem~\ref{thm:asymp}, we derived conditions that ensure $\Delta f_\y (\Dks;\Drks) =  \Delta \phi_\y\lrp{\Dks;\Drks|\s} $ and $\Delta f_{\bbP}(\Dks;\Drks)=\Delta \phi_\bbP \lrp{\Dks;\Drks|\s}$. Therefore, given that the conditions in Theorem~\ref{thm:asymp} are satisfied, $\Delta F_\Y (\rks) > \Delta f_{\bbP}(\rks) - 2\eta_N $, and the existence of a local minimum of $F_\Y(\Dks)$ within radii $\eps_k$ around $\D_k^0$, $k \in [K]$, is guaranteed with probability at least $1-e^{-\xi}$ as soon as $2\eta_N < \Delta f_\bbP (\rks) $. According to \eqref{eq:delt_p_LB}, $	\Delta \phi_\bbP \lrp{\Dks;\Drks|\s}
\geq \dfrac{s\bbE\{x^2\}}{8} \sum_{k \in [K]} \dfrac{\eps_k}{p_k}
\lrp{ \eps_k -  \eps_{k,\min}(\bar{\lambda})}$; therefore, it is sufficient to have for all $k \in [K]$:
	\begin{align*}
	\sqrt{\frac{8}{N}} L_k\eps_k \lrp{ 2\sqrt{\pi m_kp_k}
		+ \sqrt{\xi}}
	    < \frac{s\bbE\{x^2\}\eps_k\lrp{\eps_k - \eps_{k,\min}(\bar{\lambda})} }{8p_k},
	\end{align*}
which translates into $N \geq \max_{k \in [K]} N_k$, where
	\begin{align} \label{eq:N_k}
	&N_k= \lrp{ 2\sqrt{\pi m_kp_k} + \sqrt{\xi}}^2
	\lrp{ \frac{2^{4.5}L_k p_k }{s\bbE \{x^2\} (\eps_k - \eps_{k,\min}(\bar{\lambda}))}}^2 .
	\end{align}		
Furthermore, we can upper bound $L_k$ by
	\begin{align} \label{eq:Rk_UB}
	L_k &\numrel{\leq}{r_Rk} \sqrt{2}\bigg(1.25^{K/2} M_x + M_w + 2^{K/2} \lambda \sqrt{s} \bigg)^2 \nonum \\
	&\numrel{\leq}{r_lmb_Mx} \sqrt{2}c_1 \bigg(\big(1.25^{K} +  2^{K} \bar{\lambda}^2 \big) M_x^2 + M_w^2 \bigg),
	\end{align}
where $c_1$ is some positive constant, \eqref{r_Rk} follows from the fact that given the assumption in~\eqref{eq:cond_delt_s_prop}, assumptions in Lemma~\ref{lem:phi_m_T1_lip} are satisfied with $\sqrt{1-\delta_k}\geq \sqrt{1/2}$ for any $\eps_k\leq 0.15$, and \eqref{r_lmb_Mx} follows from the following inequality:
	\begin{align*}
	\lambda
		= \bar{\lambda} \bbE\lr{ |x| }
		= \dfrac{1}{s}\bar{\lambda} \bbE\lr{ \norm{\x}_1 }
		\leq \dfrac{1}{\sqrt{s}} \bar{\lambda} \bbE\lr{ \norm{\x}_2 }
		\leq\dfrac{1}{\sqrt{s}} \bar{\lambda} M_x.
	\end{align*}
Substituting \eqref{eq:Rk_UB} in \eqref{eq:N_k} and using $\lrp{ \sqrt{\xi} +  2\sqrt{\pi m_kp_k} }^2 \leq c_2 (\xi + m_kp_k)$ for some positive constant $c_2$, we get
	\begin{align*} %
	&N_k =
		\Omega \bigg(
		p_k^2 (m_kp_k+\xi)
		\bigg( \frac{2^{K}(1 + \bar{\lambda}^2) M_x^2 + M_w^2}
		{s^2\bbE\{x^2\}^2(\eps_k - \eps_{k,\min}(\bar{\lambda}))^2} \bigg) \bigg)
		\nonum \\
	&=
		\Omega \bigg(
		\frac{p_k^2 (m_kp_k+\xi) } {(\eps_k - \eps_{k,\min}(\bar{\lambda}))^2}
		\bigg( \frac{2^{K}(1 + \bar{\lambda}^2) M_x^2}
		{s^2\bbE\{x^2\}^2}
		+ \frac{M_w^2}{s^2\bbE\{x^2\}^2} \bigg) \bigg).
	\end{align*}
and $N \geq \max_{k \in [K]} N_k $.
\end{IEEEproof}

\begin{remark}
To bound deviations of $\Delta \phi_\y\lrp{\Dks;\Drks|\s}$ from its mean,
we can also use the bound provided in~\cite[Theorem 1]{gribonval2015sample} that prove uniform convergence results using covering number arguments for various classes of dictionaries. In this case, we get $\eta_N \leq c\sqrt{\dfrac{\lrp{\sum_k m_kp_k + \xi}\log N}{N}}$ for some constant $c$, where an extra $\sqrt{\log N}$ term appears compared to \eqref{eq:dev}. Therefore, Lemma~\ref{lemma:delt_m_T_dev} provides a tighter upper bound.
\end{remark}

\section{Conclusion} \label{sec:discuss}
In this paper, we focused on local recovery of coordinate dictionaries comprising a Kronecker-structured dictionary used to represent $K$th-order tensor data. We derived a sample complexity upper bound for coordinate dictionary identification up to specified errors by expanding the objective function with respect to individual coordinate dictionaries and using the coordinate-wise Lipschitz continuity property of the objective function. This analysis is local in the sense that it only guarantees existence of a local minimum of the KS-DL objective function within some neighborhood of true coordinate dictionaries. Global analysis of the KS-DL problem is left for future work.
Our results hold for dictionary coefficients generated according to the separable sparsity model. This model has some limitations compared to the random sparsity model and we leave the analysis for the random sparsity model for future work also. Another future direction of possible interest includes providing practical KS-DL algorithms that achieve the sample complexity scaling of Theorem~\ref{thm:finite_n}.

\section*{Appendix A}
\begin{IEEEproof}[Proof of Lemma~\ref{lem:Dtld}]
To prove the existence of such a formation for any $K\geq 2$, we use induction.
For $K=2$, we have
	\begin{align} \label{eq:k2}
	\lrp{\D_1\otimes \D_2} &- \lrp{\D_1^0\otimes \D_2^0} \nonum\\
	&= \lrp{\D_1 - \D_1^0} \otimes \D^0_2 + \D_1 \otimes \lrp{\D_2 - \D_2^0} \nonum \\
	&= \lrp{\D_1 - \D_1^0} \otimes \D_2 + \D^0_1 \otimes \lrp{\D_2 - \D_2^0} .
	\end{align}
For $K$ such that $K>2$, we assume the following holds:
	\begin{align} \label{eq:kK}
	&\bo_{k \in [K]} \D_k - \bo_{k \in [K]} \D_k^0 \nonum\\
	&\qquad = \sum _{k \in [K]} \wt{\D}_{k,1} \otimes \dots \otimes
		\lrp{\D_k - \D^0_k}	\otimes \dots \otimes \wt{\D}_{k,K}.
	\end{align}
Then, for $K+1$, we have:
	\begin{align} \label{eq:kKp1}
	&\bo_{k \in [K+1]} \D_k - \bo_{k \in [K+1]} \D_k^0 \nonum\\
	&= \bigg( \bo_{k \in [K]} \D_k \bigg) \otimes \D_{K+1}
		-  \bigg( \bo_{k \in [K]} \D_k^0 \bigg) \otimes \D_{K+1}^0 \nonum\\
	&\numrel{=}{r_k2} \bigg( \bo_{k \in [K]} \D_k -\bo_{k \in [K]} \D_k^0 \bigg)
		\otimes \D_{K+1}^0  \nonum\\
	&+ \bigg( \bo_{k \in [K]} \D_k \bigg) \lrp{\D_{K+1} - \D_{K+1}^0} \nonum\\
	&\numrel{=}{r_kK} \bigg( \sum _{k \in [K]} \wt{\D}_{k,1} \otimes \dots \otimes  \lrp{\D_k - \D^0_k}	\otimes \dots \otimes \wt{\D}_{k,K} \bigg)
		 \nonum \\
	&\qquad \qquad \otimes \D_{K+1}^0 + \bigg( \bo_{k \in [K]} \D_k \bigg) \lrp{\D_{K+1} - \D_{K+1}^0} \nonum\\
	&\numrel{=}{r_allcases}
	 	\sum _{k \in [K+1]} \wt{\D}_{k,1} \otimes \dots \otimes  \lrp{\D_k - \D^0_k}	\otimes \dots \otimes \wD_{k,K+1} ,
	\end{align}
where \eqref{r_k2} follows from \eqref{eq:k2}, \eqref{r_kK} follows from \eqref{eq:kK} and \eqref{r_allcases} follows from replacing $\D^0_{K+1}$ with $\wD_{k,K+1}$ in the first $K$ terms of the summation and $\D_k$'s with $\wD_{K+1,k}$, for $k \in[K]$, in the $(K+1)$th term of the summation.
\end{IEEEproof}

\begin{IEEEproof}[Proof of Lemma~\ref{lemma:f_closedform}]
Using the same definition as Gribonval et al.~\cite[Definition 1]{gribonval2014sparse}, taking the derivative of $\phi_\y\lrp{\Dks|\s}$ with respect to $\x$ and setting it to zero, we get the expression in~\eqref{eq:xhat_closedform} for $\wh{\x}$. Substituting $\wh{\x}$ in~\eqref{eq:delt_inf}, we get
 	\begin{align*}
	&\phi_\y\lrp{\Dks|\s} = \frac{1}{2}
		\bigg[
		\|\y\|_2^2 - \lrp{ \lrp{\bo \D_{k,\cJ_k}^\top} \y -\lambda \s_\cJ }^\top \nonum\\
	&\qquad \lrp{ \bo (\D_{k,\cJ_k}^\top \D_{k,\cJ_k})^{-1} }
		\lrp{ \lrp{\bo \D_{k,\cJ_k}^\top} \y -\lambda \s_\cJ }
		\bigg]\nonum\\
	&\qquad \qquad \quad  \numrel{=}{r_dP} \frac{1}{2}\|\y\|_2^2
		- \frac{1}{2} \y^\top \lrp{\bo \PDk} \y \nonum\\
	&\qquad + \lambda \s_\cJ^\top \lrp{\bo \PsDk}\y
		-\frac{\lambda^2}{2}\s_\cJ^\top \lrp{ \bo \HDk}  \s_\cJ,
	\end{align*}
where \eqref{r_dP} follows from \eqref{eq:P_H_Ps}.
\end{IEEEproof}

\begin{IEEEproof}[Proof of Lemma~\ref{lem:exp_phi}]
We use the expression for $\phi_\y\lrp{\Dks|\s}$ from~\eqref{eq:delt_closedform}. For any $\D=\bo \D_k,\D'=\bo \D'_k$, $\D_k,\D_k' \in \cD_k$, we have
	\begin{align} \label{eq:delt_ph_1}
	&\Delta \phi_\y\lrp{ \Dks;\Dkp | \s}
	= \phi_\y\lrp{\Dks|\s} - \phi_\y\lrp{\Dkp|\s} \nonum \\
	&\quad= 	\frac{1}{2} \y^\top  \lrp{ \bo \PDkp - \bo \PDk } \y \nonum\\
	&\qquad- \lambda \s_\cJ^\top \lrp{ \bo \PsDkp - \bo \PsDk } \y  \nonum \\
	&\qquad + \frac{\lambda^2}{2}\s_\cJ^\top \lrp{ \bo \HDkp - \bo \HDk } \s_\cJ.
	\end{align}
We substitute $\y = \lrp{\bo \D^0_k} \x +\n = \lrp{\bo \D^0_{k,\cJ_k} }\x_{\cJ} +\n$
and break up the sum in \eqref{eq:delt_ph_1} into 6 terms:
	\begin{align} \label{eq:delt_t}
	\Delta \phi_\y\lrp{ \Dks;\Dkp | \s} = \sum_{i\in[6]} \Delta \phi_i \lrp{ \Dks;\Dkp | \s} ,
	\end{align}	
where
	\begin{align} \label{eq:delt_phi}
	& \Delta \phi_1 \lrp{ \Dks;\Dkp | \s}
		=  \frac{1}{2} {\x}^\top \lrp{\bo \D^0_k }^\top  \nonum\\
	&\qquad	\lrp{ \bo \PDkp - \bo \PDk }
		\lrp{\bo \D^0_k } \x \nonum\\
	&\numrel{=}{r_Dtild}  \frac{1}{2} {\x}^\top  \lrp{\bo \D^0_k }^\top
		\bigg( \sum_{k \in [K]} \bP_{\wt{\D}_{1,\cJ_1}} \otimes \dots \otimes
		\nonum\\		
	&\qquad \lrp{\PDkp - \PDk}	\otimes \dots \otimes \bP_{\wt{\D}_{K,\cJ_K}}\bigg)
		\lrp{\bo \D^0_k } \x \nonum \\
	&= \frac{1}{2} {\x}^\top
		\bigg( \sum_{k \in [K]} \lrp{{\D^0_1}^\top \bP_{\wt{\D}_{1,\cJ_1}}\D^0_1} \otimes \dots \otimes  \nonum\\
	&\qquad \lrp{{\D^0_k}^\top(\PDkp - \PDk)\D^0_k}	\otimes \dots \otimes \nonum \\
	&\qquad
		\lrp{{\D^0_K}^\top \bP_{\wt{\D}_{K,\cJ_K}}\D^0_K} \bigg)
		\x , \nonum \\
	&\Delta \phi_2 \lrp{ \Dks;\Dkp | \s}
	= \n^\top
		\bigg( \sum_{k \in [K]} \lrp{\bP_{\wt{\D}_{1,\cJ_1}}\D^0_1}
		\otimes \dots \otimes \nonum\\
	&\qquad
		\lrp{(\PDkp - \PDk) \D^0_k}			
		\otimes \dots \otimes
		\lrp{ \bP_{\wt{\D}_{K,\cJ_K}} \D^0_K}\bigg)
		\x, \nonum \\
	&\Delta \phi_3 \lrp{ \Dks;\Dkp | \s}
		= \frac{1}{2} \n^\top
		\bigg( \sum_{k \in [K]} \bP_{\wt{\D}_{1,\cJ_1}} \otimes \dots \otimes \nonum\\
	&\qquad \lrp{\PDkp - \PDk}
		\otimes \dots \otimes  \bP_{\wt{\D}_{K,\cJ_K}} \bigg)
		\n, \nonum \\
	&\Delta \phi_4 \lrp{ \Dks;\Dkp | \s}
		= - \lambda \s_\cJ^\top
		\bigg( \sum_{k \in [K]} \lrp{ {\wt{\D}_{1,\cJ_1}}^+\D^0_1} \otimes \dots \otimes 	\nonum\\
	&\qquad
		\lrp{(\PsDkp - \PsDk ) \D^0_k} \otimes \dots \otimes
		\lrp{  {\wt{\D}_{K,\cJ_K}}^+ \D^0_K} \bigg)
		\x, \nonum\\
	&\Delta \phi_5 \lrp{ \Dks;\Dkp | \s}
		= - \lambda \s_\cJ^\top
		\bigg( \sum_{k \in [K]} {\wt{\D}_{1,\cJ_1}}^+ \otimes \dots \otimes \nonum\\
	&\qquad \lrp{\PsDkp - \PsDk }	\otimes \dots \otimes
		 {\wt{\D}_{K,\cJ_K}}^+ \bigg) \n, \text{and}\nonum \\
	&\Delta \phi_6 \lrp{ \Dks;\Dkp | \s}
		= \frac{\lambda^2}{2}\s_\cJ^\top
	 	\bigg( \sum_{k \in [K]} \bH_{\wt{\D}_{1,\cJ_1}} \otimes \dots \otimes \nonum\\
	 &\qquad \lrp{\HDkp - \HDk }
	 	\otimes \dots \otimes
		\bH_{\wt{\D}_{K,\cJ_K}} \bigg) \s_\cJ,
	\end{align}
where \eqref{r_Dtild} follows from Lemma~\ref{lem:Dtld} and analysis for derivation of $\lr{\Delta \phi_i\lrp{ \Dks;\Dkp | \s} }_{i=2}^6 $ are omitted due to space constraints.
Now, we set $\D' = \D^0$ and take the expectation of $\Delta \phi_\y\lrp{ \Dks;\{\D^0_k\} | \s}$ with respect to $\x$ and $\n$. Since the coefficient and noise vectors are uncorrelated,
	\begin{align*}
	\bbE \lr{ \Delta \phi_2 \lrp{\Dks;\Drks|\s}}=\bbE \lr{\Delta \phi_5 \lrp{ \Dks;\Drks | \s} } = 0.
	\end{align*}
We can restate the other terms as:
	\begin{align} \label{eq:delt_tr}
	 &\Delta \phi_1 \lrp{ \Dks;\Drks | \s}  \nonum\\
	 &\numrel{=}{r_Imk} \frac{1}{2}\tr \bigg[\x_\cJ  {\x}^\top_\cJ
		 \sum_{k \in [K]} \lrp{{\D^0_1}^\top \bP_{\wt{\D}_{1,\cJ_1}}\D^0_1} \otimes \dots \otimes  \nonum\\
	&\lrp{{\D^0_k}^\top(\I_{m_k} - \PDk)\D^0_k}	
		\otimes \dots \otimes
		\lrp{{\D^0_K}^\top \bP_{\wt{\D}_{K,\cJ_K}}\D^0_K}
		\bigg] ,\nonum \\
	& \Delta \phi_3 \lrp{ \Dks;\Drks | \s}  \nonum\\
	&= \frac{1}{2} \tr \bigg[ \n \n^\top
		\bigg( \sum_{k \in [K]} \bP_{\wt{\D}_{1,\cJ_1}} \otimes \dots \otimes
		\lrp{\PDrk -\PDk} \nonum\\
	&\qquad \otimes \dots \otimes \bP_{\wt{\D}_{K,\cJ_K}}\bigg) \bigg],
		\nonum\\
	 &\Delta \phi_4 \lrp{ \Dks;\Drks | \s}  \nonum\\
	 &\numrel{=}{r_Isk} - \lambda \tr \bigg[  \x_\cJ \s_\cJ^\top
		\bigg( \sum_{k \in [K]} \lrp{ {\wt{\D}_{1,\cJ_1}}^+\D^0_1} \otimes \dots \otimes  \nonum\\
	&\qquad \lrp{\I_{s_k} - \PsDk \D^0_k}	\otimes \dots \otimes
		\lrp{  {\wt{\D}_{K,\cJ_K}}^+ \D^0_K}\bigg)  \bigg] ,\text{and}\nonum \\
	& \Delta \phi_6 \lrp{ \Dks;\Drks | \s}  \nonum\\
	&= \frac{\lambda^2}{2} \tr \bigg[ \s_\cJ \s_\cJ^\top
	 	\bigg( \sum_{k \in [K]} \bH_{\wt{\D}_{1,\cJ_1}} \otimes \dots \otimes  \nonum\\
	&\qquad \lrp{\bH_{\D^0_{k,\cJ_k}}  - \HDk }	\otimes \dots \otimes
		\bH_{\wt{\D}_{K,\cJ_K}} \bigg) \bigg],
	\end{align}
where \eqref{r_Imk} and \eqref{r_Isk} follow from the facts that $\bP_{\D^0_{k,\cJ_k}}\D_k^0 =\D_k^0$ and ${\D^0}^+_{k,\cJ_k} \D^0_k = \I_{s_k}$, respectively. Taking the expectation of the terms in~\eqref{eq:delt_tr}, we get
	\begin{align} \label{eq:jam_T}
	& \bbE \lr{ \Delta \phi_1 \lrp{ \Dks;\Drks | \s} } \nonum \\
	&\numrel{=}{r_tr_eq}  \frac{\bbE\{x^2\}}{2}\bbE_\cJ
		\bigg\{ \sum_{k \in [K]} \tr \lrb{ {\D^0_1}^\top \bP_{\wt{\D}_{1,\cJ_1}}\D^0_1} \dots \nonum\\
	&\tr \lrb{ {\D^0_k}^\top(\I_{m_k} - \PDk)\D^0_k} \dots
		\tr \lrb{{\D^0_K}^\top \bP_{\wt{\D}_{K,\cJ_K}}\D^0_K}\bigg\}
		\nonum \\
	&=  \frac{\bbE\{x^2\}}{2}
		\sum_{k \in [K]} \bbE_{\cJ_1} \lr{ \tr \lrb{ {\D^0_1}^\top \bP_{\wt{\D}_{1,\cJ_1}}\D^0_1} }  \dots\nonum\\
	&\qquad \bbE_{\cJ_k} \lr{ \tr \lrb{  {\D^0_k}^\top(\I_{m_k} - \PDk)\D^0_k}} \dots \nonum \\
	& \qquad
		\bbE_{\cJ_K} \lr{ \tr \lrb{{\D^0_K}^\top \bP_{\wt{\D}_{K,\cJ_K}}\D^0_K}}, \nonum \\
	& \bbE \{ \Delta \phi_3 \lrp{ \Dks;\Drks | \s} \}  \nonum \\
	&= \frac{\bbE\{w^2\}}{2}
		\bbE_\cJ
		\bigg\{ \tr \bigg[\sum_{k \in [K]} \bP_{\wt{\D}_{1,\cJ_1}} \otimes \dots \otimes  \nonum\\
	&\qquad \lrp{\PDrk -\PDk} \otimes \dots \otimes \bP_{\wt{\D}_{K,\cJ_K}} \bigg] 	
		\bigg\}
		\nonum \\
	&= \frac{\bbE\{w^2\}}{2}
		\bbE_\cJ
		\bigg\{ \sum_{k \in [K]}  \tr \lrb{ \bP_{\wt{\D}_{1,J_1}} } \dots
		 \tr \lrb{ \PDrk -\PDk} \nonum\\
	&\qquad \dots  \tr \lrb{  \bP_{\wt{\D}_{K,\cJ_K}} } \bigg\} \nonum\\
	&\numrel{=}{r_proj_tr} 0, \nonum \\
	& \bbE \lr{ \Delta \phi_4 \lrp{ \Dks;\Drks | \s} } \nonum \\
	& =- \lambda \bbE\{|x|\}
		\sum_{k \in [K]} \bbE_{\cJ_1} \lr{ \tr \lrb{ {\wt{\D}_{1,\cJ_1}}^+\D^0_1}} \dots \nonum\\
	&\quad \bbE_{\cJ_k} \lr{ \tr \lrb{\I_{s_k} - \PsDk \D^0_k}} \dots
		\bbE_{\cJ_K} \lr{ \tr \lrb{  {\wt{\D}_{K,\cJ_K}}^+ \D^0_K}},
	 \nonum \\
	& \bbE \lr{ \Delta \phi_6 \lrp{ \Dks;\Drks | \s} } \nonum \\
	&= \frac{\lambda^2}{2}
		\sum_{k \in [K]} \bbE_{\cJ_1} \lr{ \tr \lrb{ \bH_{\wt{\D}_{1,\cJ_1}}} }\dots  \nonum\\
	&\quad \bbE_{\cJ_k} \lr{ \tr \lrb{ \bH_{\D^0_{k,\cJ_k}}  - \HDk }	} \dots
		\bbE_{\cJ_K} \lr{ \tr \lrb{ \bH_{\wt{\D}_{K,\cJ_K}}}} .
	\end{align}	
where \eqref{r_tr_eq} follows from the relation $\tr(\A\otimes \B)=\tr[\A]\tr[\B]$~\cite{horn2012matrix} and \eqref{r_proj_tr} follows from the fact that $\PDk$'s are orthogonal projections onto subspaces of dimension $s_k$ and $ \tr \lrb{\PDrk -\PDk}=s_k - s_k=0$.
Adding the terms in \eqref{eq:jam_T}, we obtain the expression in \eqref{eq:delt_ph_2}.
\end{IEEEproof}

\begin{IEEEproof} [Proof of Lemma~\ref{lem:UB_mu_delt_RIP}]
Equation \eqref{eq:l2_delt} follows from the definition of $\RIP$ and \eqref{eq:delt_mu} follows from Gerschgorin's disk theorem~\cite{HornJohnson,horn2012matrix,golub2012matrix}.
\end{IEEEproof}

\begin{IEEEproof} [Proof of Lemma~\ref{lem:A_B_delt}]
To lower bound $\Delta \phi_\bbP\lrp{ \Dks; \Drks | \s}$, we bound each term in~\eqref{eq:delt_ph_2} separately. For the first term $\bbE\lr{ \Delta \phi_1 \lrp{ \Dks;\Drks | \s} }$, we have
	\begin{align} \label{eq:P_Dt_LB}
	&\bbE_{\cJ_k} \lr{ \tr \lrb{ {\D^0_k}^\top \bP_{\wD_{k,\cJ_k}}\D^0_k} }	
		=  \bbE_{\cJ_k} \lr{ \norm{\bP_{\wD_{k,\cJ_k}}\D^0_{k,\cJ_k}}_F^2} .
	\end{align}
If $\wD_k = \D_k^0$, then
	\begin{align}
	\bbE_{\cJ_k} \lr{ \norm{ \PDrk \D^0_{k,\cJ_k}}_F^2} &
		\numrel{=}{r_Esp} \frac{s_k}{p_k} \norm{\D^0_k}_F^2 = s_k,
	\end{align}
where~\eqref{r_Esp} follows from~\cite[Lemma 15]{gribonval2014sparse}. If $\wD_k = \D_k$, then 	
	\begin{align*}
	& \bbE_{\cJ_k} \lr{ \norm{\bP_{\D_{k,\cJ_k}}\D^0_{k,\cJ_k}}_F^2}
	\numrel{=}{r_DC} \bbE_{\cJ_k} \lr{ \norm{ [\D_k \C_k^{-1}]_{\cJ_k}}_F^2 }  \nonum \\
	& \numrel{=}{r_exp} \frac{s_k}{p_k} \norm{\D_k\C_k^{-1}}_F^2
		\numrel{=}{r_d_j_1} \frac{s_k}{p_k}
		\sum_{j=1}^{p_k} \frac{1}{\cos^2(\theta_{(k,j)})}
		\numrel{\geq}{r_cos_ineq} \frac{s_k}{p_k} p_k = s_k,
	\end{align*}
where \eqref{r_DC} is a direct consequence of Lemma~\ref{lem:D_12_tet}; we can write $\D^0_k = \D_k\C_k^{-1} - \mb{V}_k\T_k$ where $\C_k = \Diag{\cos(\tet_{k})}$, $\T_k = \Diag{\tan(\tet_k)}$ and $\theta_{k,j}$ denotes the angle between $\bd_{k,j}$ and $\bd^0_{k,j}$. Hence $\PDk \D^0_{k,\cJ_k} =  [\D_k \C_k^{-1}]_{\cJ_k}$. Moreover,~\eqref{r_exp} follows from~\cite[Lemma 15]{gribonval2014sparse}, \eqref{r_d_j_1} follows from the fact that $\|\bd_{k,j}\|_2=1$, and \eqref{r_cos_ineq} follows from the fact that $\cos (\theta_{k,j})<1$. Similarly, we have
	\begin{align}
	& \bbE_{\cJ_k} \lr{ \tr \lrb{ {\D^0_k}^\top(\I_{m_k} - \PDk)\D^0_k} } \nonum\\
	&\qquad = \bbE_{\cJ_k} \lr{ \norm{(\I_{m_k} - \PDk)\D^0_{k,\cJ_k}}_F^2 }  \nonum \\
	&\qquad \numrel{\geq}{r_diff_p_LB} \frac{s_k}{p_k} \|\tet_k\|_2^2 \lrp{ 1 - \frac{s_k}{p_k} \frac{B^2_k}{1-\delta_k} },
	\end{align}
where~\eqref{r_diff_p_LB} follows from similar arguments as in Gribonval et al.~\cite[Equation (72)]{gribonval2014sparse}.
Putting it all together, we have
	\begin{align} \label{eq:delt1_LB}
	\bbE &\lr{ \Delta \phi_1 \lrp{ \Dks;\Drks | \s} } \nonum\\
	&\geq \frac{\bbE\{x^2\}}{2}
		\sum_{k \in [K]} \bigg(\prod_{\substack{i \in [K]\\ i \neq k}} s_i\bigg) \frac{s_k}{p_k} \|\tet_k\|_2^2
		\lrp{ 1 - \frac{s_k}{p_k} \frac{B^2_k}{1-\delta_k} } \nonum \\
	&= \frac{s \bbE\{x^2\}}{2}
		\sum_{k \in [K]} \frac{\|\tet_k\|_2^2 }{p_k}
		\lrp{ 1 - \frac{s_k}{p_k} \frac{B^2_k}{1-\delta_k} }.
	\end{align}

Next, to lower bound $\bbE\lr{ \Delta \phi_4 \lrp{ \Dks;\Drks | \s} }$, we upper bound $\lra{\bbE\lr{ \Delta \phi_4 \lrp{ \Dks;\Drks | \s} }}$. If $\wD_k = \D_k^0$, we have
	\begin{align}
	\bbE_{\cJ_k} \lr{ \tr \lrb{ {\D^0}_{k,\cJ_k}^+\D^0_{k,\cJ_k}}}
		= \bbE_{\cJ_k} \lr{ \tr \lrb{ \I_{s_k}}} = s_k,
	\end{align}
otherwise, if $\wD_k = \D_k$, we get
	\begin{align}
	&\left| \bbE_{\cJ_k} \lr{ \tr \lrb{ {\D_{k,\cJ_k}}^+\D^0_k}} \right| \nonum\\
	&\qquad \numrel{\leq}{r_tr_spn} s_k \bbE_{\cJ_k} \lr{ \norm{ {\D}_{k,\cJ_k}^+\D^0_{k,\cJ_k}}_2}
		\nonum \\
	&\qquad  \leq  s_k \bbE_{\cJ_k} \lr{ \| {\D}_{k,\cJ_k}^+\|_2\|\D^0_{k,\cJ_k} \|_2}
		\nonum \\
	&\qquad \numrel{\leq}{r_5} s_k \lrp{ \frac{1}{\sqrt{1-\delta_{s_k}(\D_k)}} }
		\lrp{ \sqrt{1+\delta_{s_k}(\D^0_k)}}
		\nonum \\
	&\qquad  \numrel{\leq}{r_delts} s_k \sqrt{\frac{1+\delta_k}{1-\delta_k}},
	\end{align}
where \eqref{r_tr_spn} follows from the fact that for a square matrix $\A \in \bbR^{q\times q}$, $\tr \lrb{\A} \leq q\|\A\|_2$, \eqref{r_5} follows from \eqref{eq:l2_delt} and \eqref{eq:pso_cond} and \eqref{r_delts} follows from \eqref{eq:A_B_Delk}. Similar to~\cite[Equation (73)]{gribonval2014sparse}, we also have
	\begin{align}
	&\left| \bbE_{\cJ_k} \lr{ \tr \big[\I_{s_k} - \PsDk \D^0_k\big]} \right|
	\nonum\\
	&\qquad \leq \frac{s_k}{p_k} \frac{\|\tet_k\|_2^2}{2}
		+ \frac{s_k^2}{p_k^2} \frac{A_kB_k}{1-\delta_k}\|\tet_k\|_2.
	\end{align}
Thus, defining $\delta_{-k} \triangleq \prod_{\substack{i \in [K]\\ i \neq k}} \sqrt{\dfrac{1+\delta_i}{1-\delta_i}}$, we get
	\begin{align}  \label{eq:delt4_LB}
	 &\bbE \lr{ \Delta \phi_4 \lrp{ \Dks;\Drks | \s} } \nonum\\
	&\quad \geq - \lambda \bbE\{|x|\}
		\sum_{k \in [K]}\delta_{-k}\bigg( \prod_{\substack{i \in [K]\\ i \neq k}}s_i
		 \bigg) \nonum\\
	&\quad \qquad \lrp{\frac{s_k}{p_k} \frac{\|\tet_k\|_2^2}{2}
		+ \frac{s_k^2}{p_k^2} \frac{A_kB_k}{1-\delta_k}\|\tet_k\|_2}
		 \nonum \\
	&\quad  = - \lambda s \bbE\{|x|\}  \sum_{k \in [K]} \frac{\delta_{-k}}{p_k}
		\lrp{ \frac{\|\tet_k\|_2^2}{2}
		+ \frac{s_k}{p_k} \frac{A_kB_k}{1-\delta_k}\|\tet_k\|_2}.
	\end{align}
	
To lower bound $\bbE\lr{ \Delta \phi_6 \lrp{ \Dks;\Drks | \s} }$, we upper bound $\lra{\bbE\lr{ \Delta \phi_6 \lrp{ \Dks;\Drks | \s} }}$. For any $\wD_k$, we have
	\begin{align}
	\lra{ \bbE_{\cJ_k} \lr{ \tr \lrb{ \bH_{\wD_{k,\cJ_k}} }} }
		&\leq \bbE_{\cJ_k} \lr{ s_k \norm{ \bH_{\wD_{k,\cJ_k}} }_2} 	
		\numrel{\leq}{r_H_UB} \frac{s_k}{1-\delta_k},
	\end{align}
where \eqref{r_H_UB} follows from \eqref{eq:pso_cond} and \eqref{eq:A_B_Delk}. Similar to Gribonval et al.~\cite[Equation (74)]{gribonval2014sparse}, we also have
	\begin{align*}
	\left| \bbE_{\cJ_k} \lr{ \tr \lrb{ \bH_{\D^0_{k,\cJ_k}} - \HDk }} \right|
	\leq \frac{s_k^2}{p_k^2} \frac{4A_kB_k}{(1-\delta_k)^2}\|\tet_k\|_2.
	\end{align*}
Thus, we get
	\begin{align}  \label{eq:delt6_LB}
	& \bbE \{ \Delta \phi_6 \lrp{ \Dks;\Drks | \s} \} \nonum\\
	&\quad \geq -  \frac{\lambda^2}{2}
		\sum_{k \in [K]} \bigg( \prod_{\substack{i \in [K]\\ i \neq k}}
		\frac{s_i}{1-\delta_i}\bigg)
		\lrp{ \frac{s_k^2}{p_k^2} \frac{4A_kB_k}{(1-\delta_k)^2}\|\tet_k\|_2}
		\nonum \\
	&\quad = - \frac{\lambda^2s}{2}
		\sum_{k \in [K]} \frac{1}{p_k}
		\bigg( \prod_{i \in [K]} \frac{1}{1-\delta_i}\bigg)
		\lrp{ \frac{s_k}{p_k} \frac{4A_kB_k}{1-\delta_k}\|\tet_k\|_2}.
	\end{align}
Adding~\eqref{eq:delt1_LB}, \eqref{eq:delt4_LB}, and \eqref{eq:delt6_LB}, we get \eqref{eq: delt_ph_3}.
\end{IEEEproof}

\begin{IEEEproof} [Proof of Proposition \ref{prop:1}]
To show that $\Delta\phi_{\bbP}\lrp{\Dks;\Drks|\s} > 0$, we use Lemma~\ref{lem:A_B_delt} and prove that the right hand side of \eqref{eq: delt_ph_3} is positive under certain conditions. First, we ensure the conditions in \eqref{eq:cond_delt_r_i} and \eqref{eq:A_B_Delk} hold for Lemma~\ref{lem:H_Dps} and Lemma~\ref{lem:A_B_delt}, respectively. We set $\delta_k = \dfrac{1}{2}$, $\delta_{s_k}(\D_k) = \dfrac{1}{2}$ and $\delta_{s_k}(\D^0_k) = \dfrac{1}{4}$, for $k \in [K]$. For $\eps_k\leq 0.15 $, this ensures:
		\begin{align}
		&\sqrt{1-\delta_{s_k}(\D_k)}
			\geq \sqrt{1-\delta_{s_k}(\D^0_k)} - \eps_k, \text{ and}
			\nonum \\
		&\max\lr{ \delta_{s_k}(\D^0_k), \delta_{s_k}(\D_k)}  \leq \delta_k,
		\end{align}
and implies $\delta_k<1$ (condition for Lemmas~\ref{lem:exp_phi} and \ref{lem:H_Ps}).
Next, we find conditions that guarantee:
	\begin{align} \label{eq:tet_2_p1}
	\frac{s_k}{p_k} \frac{B^2_k}{1-\delta_k}
		+\bar{\lambda} \kappa_x^2
		\delta_{-k}
		\numrel{=}{r_deltmk} \frac{2B^2_k s_k}{p_k}
		+\bar{\lambda} \kappa_x^2
		\lrp{3}^{(K-1)/2}
		\leq \frac{1}{2},
	\end{align}
where \eqref{r_deltmk} follows from replacing $\delta_k$ with $\dfrac{1}{2}$.
If we take $\dfrac{s_k}{p_k} \leq \dfrac{1}{8B_k^2}$ and $\bar{\lambda}\leq \dfrac{1}{8\times 3^{(K-1)/2}}$, given the fact that $\kappa_x^2\leq 1$, \eqref{eq:tet_2_p1} is satisfied.\footnote{These numbers are chosen for a simplified proof and can be modified.}
Consequently, we can restate~\eqref{eq: delt_ph_3} as
	\begin{align} \label{eq:eb_1}
	&\Delta\phi_{\bbP}\lrp{\Dks;\Drks|\s}
		\geq \frac{s \bbE\{x^2\}}{4}
		\sum_{k \in [K]} \frac{\|\tet_k\|_2 }{p_k}
		\bigg[\|\tet_k\|_2 \nonum\\
	&\qquad -8 \lrp{3^{(K-1)/2}
		+2^{(K+1)}\bar{\lambda} }
		\bar{\lambda} \kappa_x^2
		\frac{s_k}{p_k} A_kB_k\bigg].
	\end{align}
From~\cite[Proof of Proposition 2]{gribonval2014sparse}, we use the following relations:
	\begin{align} \label{eq:A0_A}
	B_k\leq B^0_k  +\eps_k \leq B^0_k +1, \quad
	A_k \leq A^0_k + 2B_k\eps_k, \quad k\in [K],
	\end{align}
where $A^0_k \triangleq \norm{{\D_k^0}^\top \D_k^0 - \I_{p_k}}_F$ and $B^0_k \triangleq  \norm{\D_k^0}_2$ and \eqref{eq:A0_A} follows from matrix norm inequalities~\cite{gribonval2014sparse}.
Defining $\gamma_k \triangleq 16 \bigg(3^{(K-1)/2}
		+2^{(K+1)}\bar{\lambda} \bigg)   \bar{\lambda}\kappa_x^2
		\dfrac{B_k^2 s_k}{p_k}$ for $k \in [K]$ and using $\kappa_x^2 \leq 1$, we have
	\begin{align} \label{eq:gam}
	\gamma_k &\leq
		2\bigg(3^{(K-1)/2}
		+\frac{2^{(K+1)}}{8\times 3^{(K-1)/2}} \bigg)
		\bigg(\frac{1}{8\times 3^{(K-1)/2}} \bigg) \nonum\\
		&\leq 2\lrp{\frac{1}{8}+\frac{4}{64}}
		\leq\frac{1}{2}.
	\end{align}
Then, for $\D_k \in \cS_{\eps_k}(\D_k^0)$, $k \in [K]$, we get
	\begin{align}
	&\Delta\phi_{\bbP}\lrp{\Dks;\Drks|\s}  \nonum\\
	&\qquad \numrel{\geq}{r_tet_rk} \frac{s\bbE\{x^2\}}{4}
		\sum_{k \in [K]} \frac{\eps_k}{p_k}
		\lrp{ \eps_k - \frac{\gamma_k}{2} \frac{A_k}{B_k}	}
		\nonum \\
	&\qquad	\numrel{\geq}{r_A_B} \frac{s\bbE\{x^2\}}{4}
		\sum_{k \in [K]} \frac{\eps_k}{p_k}
		\lrp{ \eps_k - \frac{\gamma_k}{2} \frac{A_k^0+2B_k\eps_k}{B_k}	}
		\nonum \\
	&\qquad	\geq \frac{s\bbE\{x^2\}}{4}
		\sum_{k \in [K]} \frac{\eps_k}{p_k}
		\lrp{ \eps_k(1-\gamma_k) - \frac{\gamma_k}{2} \frac{A_k^0}{B_k}	}
		\nonum \\
	&\qquad	\numrel{\geq}{r_gam} \frac{s\bbE\{x^2\}}{8}
		\sum_{k \in [K]} \frac{\eps_k}{p_k}
		\lrp{ \eps_k - \gamma_k \frac{A_k^0}{B_k}	},
	\end{align}
where \eqref{r_tet_rk} follows from \eqref{eq:eb_1}, \eqref{r_A_B} follows from \eqref{eq:A0_A}, and \eqref{r_gam} follows from \eqref{eq:gam}.
Hence, we can write
	\begin{align} \label{eq:Delt_LB_rmin}
	\Delta\phi_{\bbP}\lrp{\Dks;\Drks|\s} \geq& \frac{s\bbE\{x^2\}}{8}
		\sum_{k \in [K]} \frac{\eps_k}{p_k}
		\lrp{ \eps_k -  \eps_{k,\min}(\bar{\lambda})},
	\end{align}
where we define
	\begin{align}
	& \eps_{k,\min} (\bar{\lambda})
	\triangleq  \gamma_k \frac{A_k^0}{B_k} \nonum\\
	&= 16 \lrp{3^{(K-1)/2}
		+2^{(K+1)}\bar{\lambda} }
		\bar{\lambda}\kappa_x^2
		\frac{s_k}{p_k}  A_k^0 B_k \nonum \\
	&=\frac{2}{3^{(K+1)/2}} \bigg(3^{(K-1)/2}
		+2^{(K+1)}\bar{\lambda} \bigg)  \bar{\lambda} C_{k,\min},
	\end{align}
and $C_{k,\min}$ is defined in \eqref{eq:cond_C_min_C_max}.
The lower bound in \eqref{eq:Delt_LB_rmin} holds for any $\eps_k \leq 0.15$ and $\D_k \in \cS_{\eps_k}(\D_k^0)$, $k \in [K]$.
Finally, since $3^{(K-1)/2} +2^{(K+1)}\bar{\lambda} \leq 0.5\times 3^{(K+1)/2}$,
the assumption
$\bar{\lambda} \leq 0.15/(\max_{k \in [K]} C_{k,\min})$
implies that  $\eps_{k,\min}(\bar{\lambda}) \leq 0.15$ for $k\in[K]$.
Therefore, $\Delta\phi_{\bbP}\lrp{\Dks;\Drks|\s} > 0 $ for all $\eps_k \in (\eps_{k,\min}(\bar{\lambda}),0.15]$, $k \in [K]$.
\end{IEEEproof}

\begin{IEEEproof}[Proof of Lemma~\ref{lem:mu_mu0_rel}]
Considering $j \not \in \cJ$, associated with $\lrp{j_1,\dots,j_k} \not \in \lrp{\cJ_1\times \dots \times \cJ_K}$, we have
	\begin{align}
	&\|\D_\cJ^\top \bd_j\|_1  \nonum\\
	&\numrel{\leq}{r_tr_ineq} \|{\D_\cJ^0}^\top \bd^0_j\|_1 + \|{\D_\cJ^0}^\top (\bd_j-\bd^0_j)\|_1
		+ \|(\D_\cJ-\D_\cJ^0)^\top \bd_j\|_1\nonum \\
	&\leq \mu_s(\D^0)
		+ \sqrt{s} \lrb{ \|{\D_\cJ^0}^\top( \bd_{j}-\bd_{j}^0) \|_2
		+ \|(\D_\cJ-\D^0_\cJ)^\top\bd_j\|_2} \nonum \\
	&\leq \mu_s(\D^0) + \sqrt{s}
		\bigg[ \norm{ \bo {\D^0_{k,\cJ_k}}^\top}_2
		\norm{ \bo \lrp{ \bd_{k,j_k} - \bd^0_{k,j_k} }}_2 \nonum\\
	&\qquad + \norm{\bo \D_{k,\cJ_k} -\bo \D^0_{k,\cJ_k}}_2
		 \norm{\bd_j}_2 \bigg] \nonum \\
	&\numrel{\leq}{r_d0_mu} \mu_s(\D^0) + \sqrt{s}
		\bigg[ \bigg( \prod_{k \in [K]} \sqrt{1+\delta_{s_k}(\D^0_k)} \bigg)
		\nonum\\
	&\qquad \bigg( \sum_{k \in [K]} \norm{ \wt{\bd}_{1,j_1}}_2\dots
		\norm{\bd_{k,j_k} - \bd^0_{k,j_k} }_2 \dots \
		\norm{ \wt{\bd}_{k,j_K}}_2 \bigg)  \nonum \\
	 &\quad
	 	+ \sum_{k \in [K]} \norm{ \wD_{1,\cJ_1}}_2\dots
		\norm{\D_{k,\cJ_k} - \D^0_{k,\cJ_k} }_2 \dots \
		\norm{ \wD_{k,\cJ_k}}_2 \bigg] \nonum \\
	&\numrel{\leq}{r_RIP_d} \mu_s(\D^0) + \sqrt{s}
		\bigg[ \bigg( \prod_{k \in [K]} \sqrt{1+\delta_{s_k}(\D^0_k)} \bigg)
		\bigg( \sum_{k \in [K]} \eps_k \bigg) \nonum\\
	&\qquad + \sum_{k \in [K]} \bigg( \prod_{\substack{i \in [K] \\ i \neq k}}
		\norm{\wD_{i,\cJ_i}}_2 \bigg) \eps_k
		 \bigg] \nonum \\
	&\numrel{\leq}{r_asump} \mu_s(\D^0) + 2(1.5)^{K/2}\sqrt{s}  \bigg( \sum_{k \in [K]} \eps_k \bigg),
	\end{align}	
where \eqref{r_tr_ineq} follows from the triangle inequality, \eqref{r_d0_mu} follows from \eqref{eq:otD_otDp}, \eqref{r_RIP_d} follows from \eqref{eq:delt_mu}, and, \eqref{r_asump} follows from substituting the upper bound value from \eqref{eq:UB_mu_delt} for $\delta_{s_k}(\D_k^0) $. For  $\wD_i = \D_i^0$, $\norm{\D_{i,\cJ_i}^0}_2 \leq \sqrt{1 + \delta_{s_i}(\D^0_i)}  \leq \sqrt{\frac{5}{4}} < 1.5$ and for $\wD_i = \D_i$, according to \eqref{eq:A0_A}, we have $\norm{\D_{i,\cJ_i}}_2 \leq \norm{\D_{i,\cJ_i}^0}_2 + \eps_i \leq \sqrt{\frac{5}{4}}+0.15 < 1.5$.
\end{IEEEproof}

\begin{IEEEproof}[Proof of Proposition~\ref{prop:3}]
We follow a similar approach to Gribonval et al.~\cite{gribonval2014sparse}. We show that the conditions in~\eqref{eq:lem8_cod} hold for Lemma~\ref{lem:a_hat_min_cond}. We have
	\begin{align}
	& \norm{\y - \lrp{ \bo \D_k} \x }_2 \nonum\\
	&\leq \norm{ \lrp{ \bo \D^0_{k,\cJ_k} - \bo \D_{k,\cJ_k} }
		\x_\cJ} _2+\|\n\|_2 \nonum \\
	&\leq M_x
		\sum _{k \in [K]} \big\| \wD_{1,\cJ_1} \otimes \dots \otimes
		\lrp{\D^0_{k,\cJ_k} - \D_{k,\cJ_k} }	\otimes \dots \otimes \nonum\\
	&\qquad  \wD_{K,\cJ_K}\big\|_2  +M_w \nonum \\
	&\leq M_x
		\sum _{k \in [K]} \norm{ \wD_{1,\cJ_1}}_2 \dots
		\norm{\D^0_{k,\cJ_k} - \D_{k,\cJ_k} }_2 \dots \norm{ \wD_{K,\cJ_K}}_2  \nonum\\
	&\qquad +M_w
		\nonum \\
	&\leq M_x
		\sum _{k \in [K]}
		\bigg( \prod_{\substack{i \in [K] \\ i \neq k}}
		\norm{ \wD_{i,\cJ_i }}_2 \bigg) \eps_k
		 +M_w \nonum \\
	&\numrel{\leq}{r_delt_leq} (1.5)^{(K-1)/2} M_x  \sum_{k \in [K]} \eps_k  + M_w,
	\end{align}
where \eqref{r_delt_leq} follows from \eqref{eq:delt_cond} and the fact that for  $\wD_i = \D_i^0$, $\norm{\D_{i,\cJ_i}^0}_2 \leq \sqrt{1 + \delta_{s_i}(\D^0_i)}  \leq \sqrt{\frac{5}{4}} < 1.5$ and for $\wD_i = \D_i$, according to \eqref{eq:A0_A}, we have $\norm{\D_{i,\cJ_i}}_2 \leq \norm{\D_{i,\cJ_i}^0}_2 + \eps_i \leq \sqrt{\frac{5}{4}}+0.15 < 1.5$.
Hence, we get
	\begin{align} \label{eq:lmb_m_err}
	&\lambda (1-2\mu_s(\D))- \norm{\y - \lrp{ \bo \D_k} \x}_2 \nonum \\
	&\geq \lambda (1-2\mu_s(\D)) -  (1.5)^{(K-1)/2} M_x  \sum_{k \in [K]} \eps_k  -M_w
		\nonum \\
	&\numrel{\geq}{r_mu_mu0} \lambda (1-2\mu_s(\D^0))
		- (1.5)^{K/2} \lrp{ 4\lambda \sqrt{s} + (1.5)^{-1/2}M_x} \nonum\\
	&\qquad \sum_{k \in [K]} \eps_k
		-M_w \nonum \\
	&\numrel{\geq}{r_lam_m_a}  \lambda (1-2\mu_s(\D^0)) - 3(1.5)^{K/2} M_x \sum_{k \in [K]} \eps_k -M_w
	\nonum \\
	&=  3(1.5)^{K/2}  M_x \bigg( K \bar{\lambda} C_{\max} -  \sum_{k \in [K]} \eps_k  \bigg) -M_w,
	\end{align}
where \eqref{r_mu_mu0} follows from \eqref{eq:mu_mu0} and \eqref{r_lam_m_a} follows from \eqref{eq:lem8_cod} ($2\lambda \sqrt{s} \leq x_{\min}\sqrt{s}\leq M_x$) and~\eqref{eq:mu_mu0}.
If $\eps_k < C_{\max} \bar{\lambda} $, $k \in [K]$, the assumption on the noise level in \eqref{eq:M_eps_M_al} implies that the right-hand side of \eqref{eq:lmb_m_err} is greater than zero and $\lambda (1-2\mu_s(\D))>  \norm{\y - \lrp{ \bo \D_k} \x}_2 $. Thus, according to Lemma~\ref{lem:a_hat_min_cond}, $\wh{\x}$ is almost surely the unique solution of $\min_{\x } \frac{1}{2}\norm{\y - \lrp{ \bo \D_k} \x' }_2  +\lambda\|\x'\|_1$ and $\Delta\phi_{\bbP}\lrp{\Dks,\Drks|\s} = \Delta f_{\bbP} \lrp{\Dks,\Drks}$.
\end{IEEEproof}

\section*{appendix B}

\begin{IEEEproof} [Proof of Lemma~\ref{lemma:delt_m_T_dev}]
According to Lemma~\ref{lem:rad}, we have to upper bound
$
	\bbE \lr{ \sup_{\D_k \in \overline{\mc{B}}_{\eps_k}(\D^0_k), k \in [K]}\lra{ \frac{1}{N}
\sum_{n \in [N] } \beta_n h_n(\Dks) }}
$. Conditioned on the draw of functions $h_1,\dots,h_N$, consider the Gaussian processes
$
	A_{\Dks} = \frac{1}{N} \sum_{n \in [N]} \beta_n h_n(\Dks)
$ and
$
	C_{\Dks} = \sqrt{\frac{K}{N}}
		\sum_{k \in [K]} \bigg( L_k \sum_{i \in [m_k]} \sum_{j \in [p_k]}
		\zeta_{ij}^k  (\D_k-\D_k^0)_{ij} \bigg)
$,
where $\lr{\beta_n}_{n=1}^N$'s and $\lr{\zeta_{ij}^k},k\in [K],i\in [m_k],j\in [p_k]$'s are independent standard Gaussian vectors.
We have
	\begin{align}
	&\bbE \lr{\lra{A_{\Dks} - A_{\Dkp}}^2} \nonum\\
	&\qquad= \frac{1}{N^2} \bigg|
		\sum_{n \in [N]} h_n(\Dks)- h_n(\Dkp) \bigg|^2 \nonum \\
	&\qquad \numrel{\leq}{r_lpsh_h} \frac{1}{N}
		\bigg(\sum_{k \in [K]} L_k\FnDk \bigg)^2\nonum \\
	&\qquad \numrel{\leq}{r_cs} \frac{K}{N}\sum_{k \in [K]} L_k^2\FnDk^2\nonum \\
	&\qquad = \bbE\lr{\lra{C_{\Dks} - C_{\Dkp}}^2},
	\end{align}
where \eqref{r_lpsh_h} follows from coordinate-wise Lipschitz continuity of $h$ and \eqref{r_cs} follows from Cauchy-Schwartz inequality. Hence, using Slepian's Lemma~\cite{massart2007concentration}, we get
	\begin{align}
	\bbE \bigg\{ \sup_{\substack{\D_k \in \overline{\mc{B}}_{\eps_k}(\D^0_k)  \\ k \in [K]}}
		A_{\Dks} \bigg\}
	&\leq \bbE \bigg\{ \sup_{\substack{ \D_k \in \overline{\mc{B}}_{\eps_k}(\D^0_k)  \\ k \in [K]}}
		C_{\Dks} \bigg\} \nonum \\
	& = \sqrt{\frac{K}{N}} \bigg(\sum_{k \in [K]}  L_k\eps_k \bbE \big\{\|\boldsymbol{\zeta}^k\|_F \big\}\bigg)\nonum \\
	& =\sqrt{\frac{K}{N}} \bigg(\sum_{k \in [K]} L_k\eps_k \sqrt{m_kp_k}\bigg).
	\end{align}
Thus, we obtain
$
	\bbE \lr{\sup_{\substack{ \D_k \in \overline{\mc{B}}_{\eps_k}(\D^0_k)  \\ k \in [K]}}
		\lra{\frac{1}{N} \sum_{n \in [N]} \beta_nh_n(\Dks) } }
	\\ \leq 2\sqrt{\frac{K}{N}}\lrp{\sum_{k \in [K]} L_k\eps_k \sqrt{m_kp_k}}.
$
\end{IEEEproof}

\begin{IEEEproof}[Proof of Lemma~\ref{lem:phi_m_T1_lip}]
We expand $\Delta \phi_\y \lrp{\Dks;\Drks|\s}$ according to \eqref{eq:delt_t} and bound each term of the sum separately. Looking at the first term, we get
	\begin{align}
	& \lra{ \Delta \phi_1 \lrp{\Dks;\Drks|\s} }
		\numrel{=}{r_exp_phi1} \bigg| \frac{1}{2}\x^\top
		{\D^0 }^\top
		\bigg( \sum_{k \in [K]} \bP_{\wt{\D}_{1,\cJ_1}}
		\otimes \dots \otimes  \nonum\\
	&\qquad
		\lrp{\PDkp - \PDk}	\otimes \dots \otimes \bP_{\wt{\D}_{K,\cJ_K}}\bigg)
		\D^0
		\x \bigg| \nonum \\
	&\numrel{\leq}{r_D0_D0k} \frac{1}{2} \norm{\x}_2^2
		\bigg(\prod_{k \in [K]} \norm{ \D^0_{k,\cJ_k} }_2^2 \bigg)
		\bigg( \sum_{k \in [K]}
		\norm{\PDrk - \PDk}_2 \nonum\\
	&\qquad \bigg( \prod_{\substack{i\in [K] \\ i \neq k}}
		\norm{\bP_{\wD_{i,\cJ_i}}}_2  \bigg)\bigg)
		\nonum \\
	&\numrel{\leq}{r_p_rip1} M_x^2
		\bigg( \prod_{k \in [K]}\big(1+\delta_{s_k}(\D^0_k)\big) \bigg)
		\nonum\\
	&\qquad \bigg( \sum_{k \in [K]}(1 - \delta_k)^{-1/2}\FrDk \bigg) ,
	\end{align}
where \eqref{r_exp_phi1} follows from \eqref{eq:delt_phi}, \eqref{r_D0_D0k} follows from the fact that $\norm{\D^0_\cJ}_2 = \prod_{k \in [K]} \norm{ \D^0_{k,\cJ_k} }_2$, and \eqref{r_p_rip1} follows from  the definition of $\RIP$, equation \eqref{eq:PH_PHp}, and $\big\|\bP_{\wD_{i,\cJ_i}}\big\|_2=1$. Following a similar approach and expanding the rest of the terms, we get
	\begin{align*}
	&\lra{ \Delta \phi_2 \lrp{\Dks;\Drks|\s} } \nonum\\
	&\leq \norm{\n}_2 \norm{\x}_2
		\bigg( \prod_{k \in [K]} \norm{ \D^0_{k,\cJ_k} }_2^2 \bigg)
		\nonum\\
	&\qquad \bigg( \sum_{k \in [K]} \norm{\PDrk - \PDk }_2
	\bigg( \prod_{\substack{i\in [K] \\ i \neq k}} \norm{\bP_{\wD_{i,\cJ_i}}}_2  \bigg)\bigg)
	\nonum \\
	&\numrel{\leq}{r_p_rip} 2M_w M_x
		\bigg( \prod_{k \in [K]}\big(1+\delta_{s_k}(\D^0_k)\big)^{1/2} \bigg)
		\nonum\\
	&\qquad \bigg( \sum_{k \in [K]} (1 - \delta_k)^{-1/2}\FrDk \bigg), \nonum \\
	&\lra{ \Delta \phi_3 \lrp{\Dks;\Drks|\s} }
		\leq  \frac{1}{2} \norm{\n}_2^2 \nonum\\
	& \qquad \bigg( \sum_{k \in [K]}
		\norm{\PDrk - \PDk}_2	
		\bigg( \prod_{\substack{i\in [K] \\ i \neq k}} \norm{\bP_{\wD_{i,\cJ_i}}}_2  \bigg)
		\bigg) \nonum \\
	&\leq M_w^2 \bigg( \sum_{k \in [K]} (1 - \delta_k)^{-1/2}\FrDk \bigg),\nonum \\
	&\lra{ \Delta \phi_4 \lrp{\Dks;\Drks|\s} }
		= \lambda \norm{\s_\cJ}_2 \norm{\x}_2
		\bigg( \prod_{k \in [K]} \norm{\D^0_{\cJ_k}}_2 \bigg)
		\nonum\\
	&\qquad \bigg( \sum_{k \in [K]}
		\norm{\PsDrk - \PsDk}_2
		\bigg( \prod_{\substack{i\in [K] \\ i \neq k}}  \norm{ {\wD_{i,\cJ_i}}^+}_2 \bigg)
		 \bigg) \nonum \\
	&\numrel{\leq}{r_ps} 2\lambda \sqrt{s} M_x
		 \bigg( \prod_{k \in [K]}\big(1+\delta_{s_k}(\D^0_k)\big)^{1/2} \bigg) \nonum\\
	&\qquad \bigg( \sum_{k \in [K]} (1-\delta_k)^{-1}
		 \bigg( \prod_{\substack{i \in [K] \\ i \neq k}}
		 (1-\delta_i)^{-1/2} \bigg) \FrDk \bigg) ,\nonum \\
	& \lra{ \Delta \phi_5 \lrp{\Dks;\Drks|\s} }
	= \lambda \norm{\s_\cJ}_2 \norm{\n}_2 \nonum\\
	&\qquad \bigg( \sum_{k \in [K]}
		\norm{\PsDrk - \PsDk }_2	
		\bigg( \prod_{\substack{i\in [K] \\ i \neq k}}  \norm{ {\wD_{i,\cJ_i}}^+}_2 \bigg)
		\bigg)	\nonum \\
	&\leq 2\lambda \sqrt{s}M_w \nonum\\
	&\qquad \bigg( \sum_{k \in [K]} (1-\delta_k)^{-1}
		 \bigg( \prod_{\substack{i \in [K] \\ i \neq k}}
		 (1-\delta_i)^{-1/2} \bigg)  \FrDk \bigg) ,\nonum \\
	&\lra{ \Delta \phi_6 \lrp{\Dks;\Drks|\s} }
		= \frac{\lambda^2}{2} \norm{\s_\cJ}_2^2  \nonum \\
	&\qquad
	 	\bigg( \sum_{k \in [K]}
		\norm{\HDrk - \HDk }_2	
		\bigg( \prod_{\substack{i\in [K] \\ i \neq k}}  \norm{\bH_{\wD_{i,\cJ_i}} }_2 \bigg)\bigg)	
		\nonum \\
	&\numrel{\leq}{r_h} \lambda^2 s
		\bigg( \sum_{k \in [K]} (1-\delta_k)^{-\frac{3}{2}}
		 \bigg( \prod_{\substack{i \in [K] \\ i \neq k}}
		 (1-\delta_i)^{-1} \bigg) \FrDk\bigg),
	\end{align*}
where \eqref{r_ps} and \eqref{r_h} follow from \eqref{eq:pso_cond} and \eqref{eq:PH_PHp}. Adding all the terms together, we get
	\begin{align}
	&\lra{\Delta \phi_\y\lrp{\Dks;\Drks|\s}}  \leq
		\sum_{k \in [K]} L_k \FrDk.
	\end{align}
where $L_k$ is defined in \eqref{eq:lipsch_const_h}.
\end{IEEEproof}	
\section*{appendix C}
\begin{IEEEproof}[Proof of the coherence relation for KS dictionaries]
To prove \eqref{eq:mu_s}, we define the set $\mathcal{A} = \lr{\forall j_k \in \cJ_k, (j_1,\dots,j_K) \not\in (\cJ_1,\dots,\cJ_K)}$. We have
	\begin{align}
	\mu_s(\D)&= \max_{|\cJ|\leq s} \max_{j \not\in \cJ}\|\D_\cJ^\top \bd_j\|_1 \nonum \\
	&= \max_{\substack{|\cJ_k|\leq s_k\\ k \in [K]}}
		\max_{\mathcal{A}}
		\norm{\lrp{\bo \D_{k,\cJ_k}^\top}\lrp{\bo \bd_{k,j_k}}}_1\nonum \\
	&= \max_{\substack{|\cJ_k|\leq s_k\\ k \in [K]}}
		\max_{\mathcal{A}}
		\norm{\bo \D_{k,\cJ_k}^\top\bd_{k,j_k}}_1 \nonum\\
	&=  \max_{\substack{|\cJ_k|\leq s_k\\ k \in [K]}}
		\max_{\mathcal{A}}
		\prod_{k \in [K]} \norm{\D_{k,\cJ_k}^\top\bd_{k,j_k}}_1  \nonum\\
	&\leq \max_{k \in [K]} \mu_{s_k}(\D_k)
		\bigg( \prod_{\substack{i \in [K], \\ i \neq k}}  \lrp{ 1+\mu_{s_i-1}(\D_i)} \bigg).
	\end{align}	

\end{IEEEproof}
\bibliographystyle{IEEEtran}
\bibliography{IEEEabrv,refs}

\begin{IEEEbiography}[{\includegraphics[width=1in,height=1.25in,clip,keepaspectratio]{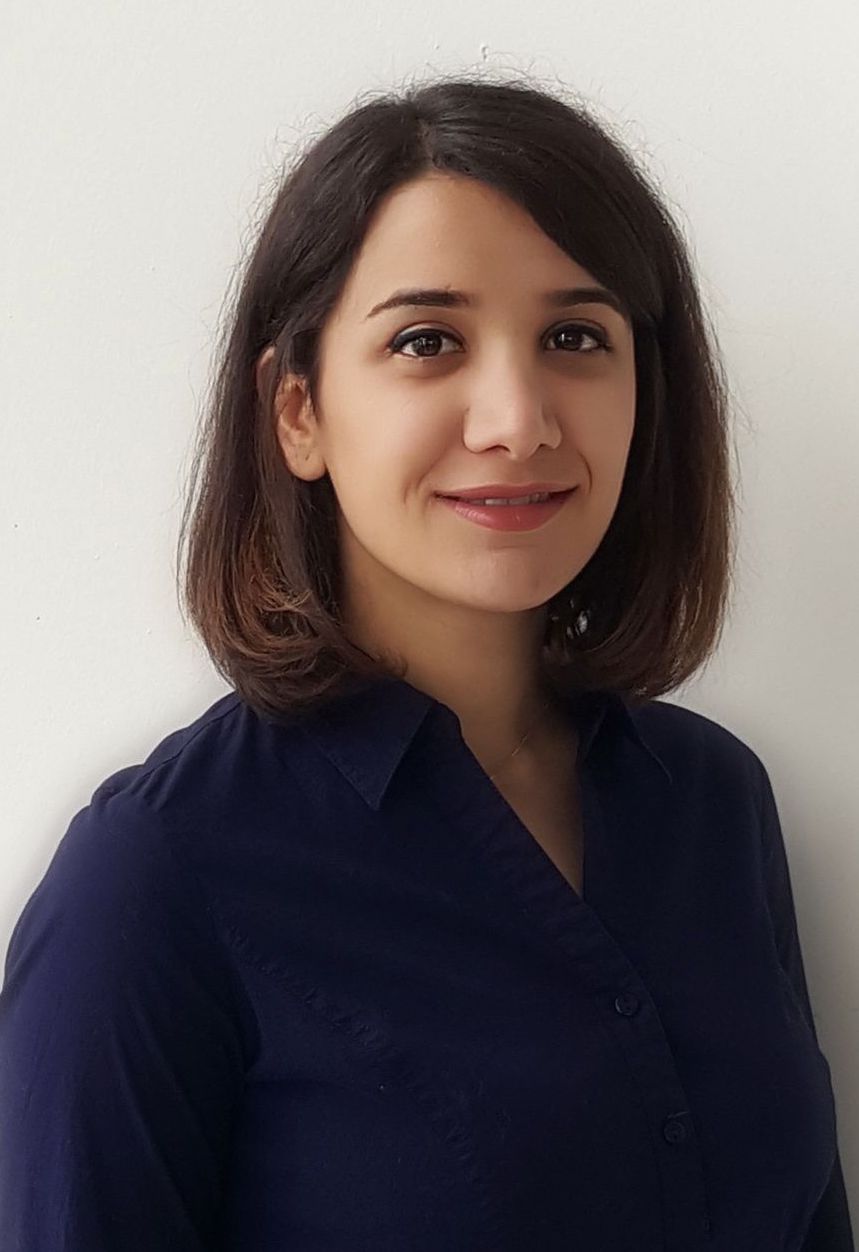}}]
{Zahra Shakeri} is pursuing a Ph.D. degree at Rutgers University, NJ, USA. She is a member of the INSPIRE laboratory. She received her M.Sc. degree in Electrical and Computer Engineering from Rutgers University, NJ, USA, in 2016 and her B.Sc. degree in Electrical Engineering from Sharif University of Technology, Tehran, Iran, in 2013.
Her research interests are in the areas of machine learning, statistical signal processing, and multidimensional data processing.
\end{IEEEbiography}

\begin{IEEEbiography}[{\includegraphics[width=1in,height=1.25in,clip,keepaspectratio]{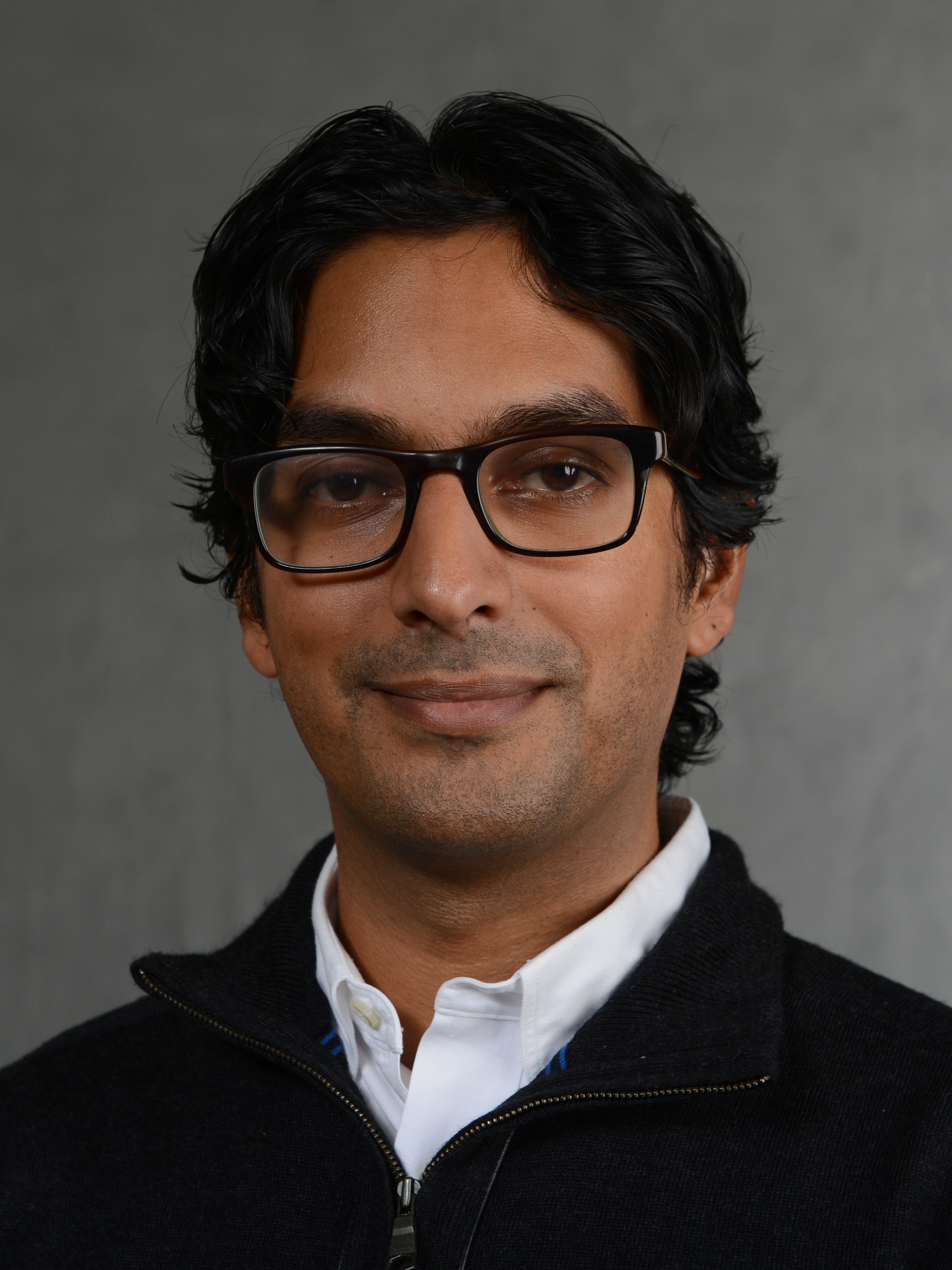}}]
{Anand D. Sarwate}(S'99--M'09--SM'14) received the B.S. degrees in electrical engineering and computer science and mathematics from the Massachusetts Institute of Technology, Cambridge, MA, USA, in 2002, and the M.S. and Ph.D. degrees in electrical engineering from the Department of Electrical Engineering and Computer Sciences (EECS), University of California, Berkeley (U.C. Berkeley), Berkeley, CA, USA. He is a currently an Assistant Professor with the Department of Electrical and Computer Engineering, The State University of New Jersey, New Brunswick, NJ, USA, since January 2014. He was previously a Research Assistant Professor from 2011 to 2013 with the Toyota Technological Institute at Chicago; prior to this, he was a Postdoctoral Researcher from 2008 to 2011 with the University of California, San Diego, CA. His research interests include information theory, machine learning, signal processing, optimization, and privacy and security. Dr. Sarwate received the A. Walter Tyson Assistant Professor Award from the Rutgers School of Engineering, the NSF CAREER award in 2015, and the Samuel Silver Memorial Scholarship Award and the Demetri Angelakos Memorial Award from the EECS Department at U.C. Berkeley. He was awarded the National Defense Science and Engineering Graduate Fellowship from 2002 to 2005. He is a member of Phi Beta Kappa and Eta Kappa Nu.
\end{IEEEbiography}

\begin{IEEEbiography}[{\includegraphics[width=1in,height=1.25in,clip,keepaspectratio]{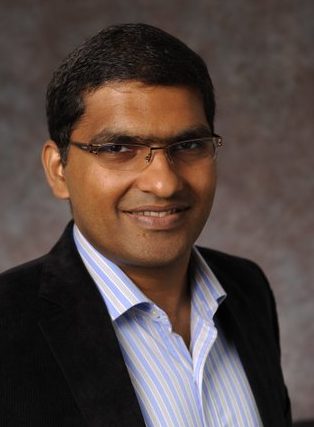}}]
{Waheed U. Bajwa} received BE (with Honors) degree in electrical engineering from the National University of Sciences and Technology, Pakistan in 2001, and MS and PhD degrees in electrical engineering from the University of Wisconsin-Madison in 2005 and 2009, respectively. He was a Postdoctoral Research Associate in the Program in Applied and Computational Mathematics at Princeton University from 2009 to 2010, and a Research Scientist in the Department of Electrical and Computer Engineering at Duke University from 2010 to 2011. He has been with Rutgers University since 2011, where he is currently an associate professor in the Department of Electrical and Computer Engineering and an associate member of the graduate faculty of the Department of Statistics and Biostatistics. His research interests include statistical signal processing, high-dimensional statistics, machine learning, harmonic analysis, inverse problems, and networked systems.

Dr. Bajwa has received a number of awards in his career including the Best in Academics Gold Medal and President's Gold Medal in Electrical Engineering from the National University of Sciences and Technology (2001), the Morgridge Distinguished Graduate Fellowship from the University of Wisconsin-Madison (2003), the Army Research Office Young Investigator Award (2014), the National Science Foundation CAREER Award (2015), Rutgers University's Presidential Merit Award (2016), Rutgers Engineering Governing Council ECE Professor of the Year Award (2016, 2017), and Rutgers University's Presidential Fellowship for Teaching Excellence (2017). He is a co-investigator on the work that received the Cancer Institute of New Jersey's Gallo Award for Scientific Excellence in 2017, a co-author on papers that received Best Student Paper Awards at IEEE IVMSP 2016 and IEEE CAMSAP 2017 workshops, and a Member of the Class of 2015 National Academy of Engineering Frontiers of Engineering Education Symposium. He served as an Associate Editor of the IEEE Signal Processing Letters (2014 – 2017), co-guest edited a special issue of Elsevier Physical Communication Journal on ``Compressive Sensing in Communications" (2012), co-chaired CPSWeek 2013 Workshop on Signal Processing Advances in Sensor Networks and IEEE GlobalSIP 2013 Symposium on New Sensing and Statistical Inference Methods, and served as the Publicity and Publications Chair of IEEE CAMSAP 2015 and General Chair of the 2017 DIMACS Workshop on Distributed Optimization, Information Processing, and Learning. He is currently serving as Technical Co-Chair of the IEEE SPAWC 2018 Workshop, Senior Area Editor for IEEE Signal Processing Letters, Associate Editor for IEEE Transactions on Signal and Information Processing over Networks, is a Senior Member of the IEEE, and serves on the MLSP, SAM, and SPCOM Technical Committees of the IEEE Signal Processing Society.
\end{IEEEbiography}

\end{document}